\documentclass[twocolumn,10pt]{article} 

\usepackage[utf8]{inputenc}  %document encoding
\usepackage[english]{babel}   %english line breaks etc.
\usepackage{lmodern}  %fonts in arbitratry sizes (removes "missing font" warning) -> some are still remaingin from the ipa package
\usepackage{microtype}  %prevents unnessesary 'overful hbox' warnings
\usepackage{tipa}		%IPA characters
\usepackage{textcomp,gensymb} %additional math symbols (\degree)
\usepackage{booktabs}   %nice tables
\usepackage{rotating}   %sideways tables
\usepackage[usenames,dvipsnames,svgnames,table]{xcolor} %colour names
\usepackage{authblk} %affiliation

\begin{document}

\title{A three-dimensional approach to Visual Speech Recognition using Discrete Cosine Transforms}
\author[1]{Toni Heidenreich\thanks{toni@heidenreich89.de}}
\author[1]{Michael W. Spratling}
\affil[1]{Department of Informatics, King's College London}
\date{August 12, 2014}

\maketitle

\abstract{ \small
Visual speech recognition aims to identify the sequence of phonemes from continuous speech. Unlike the traditional approach of using 2D image feature extraction methods to derive features of each video frame separately, this paper proposes a new approach using a 3D (spatio-temporal) Discrete Cosine Transform to extract features of each feasible sub-sequence of an input video which are subsequently classified individually using Support Vector Machines and combined to find the most likely phoneme sequence using a tailor-made Hidden Markov Model. The algorithm is trained and tested on the VidTimit database to recognise sequences of phonemes as well as visemes (visual speech units). Furthermore, the system is extended with the training on phoneme or viseme pairs (biphones) to counteract the human speech ambiguity of co-articulation. The test set accuracy for the recognition of phoneme sequences is 20\%, and the accuracy of viseme sequences is 39\%. Both results improve the best values reported in other papers by approximately 2\%. The contribution of the result is three-fold: Firstly, this paper is the first to show that 3D feature extraction methods can be applied to continuous sequence recognition tasks despite the unknown start positions and durations of each phoneme. Secondly, the result confirms that 3D feature extraction methods improve the accuracy compared to 2D features extraction methods. Thirdly, the paper is the first to specifically compare an otherwise identical method with and without using biphones, verifying that the usage of biphones has a positive impact on the result.

}

\section{Introduction}

People who claim to have the ability to know what others say by just watching their mouth movements have always enthralled onlookers. However, few human beings seem to be able to reach high recognition rates \cite{ONeill1951}. This acclaimed challenge of the task gives a clear motivation to achieve comparable results with a computer algorithm. Although researchers started to investigate artificial lip-reading as early as 1984 \cite{Petajan1984}, it continues to be of high research interest with many improvements in recent years as big datasets become available and computers provide increasingly fast computation and sufficiently large working memory. In the research context, the area is now known by different names including lip reading, speech reading and visual speech recognition. Given possible application areas, like support systems for deaf people or security and surveillance systems, improvements in the recognition rates of visual speech reading systems are desirable. Parts of lip-reading systems can also be useful for biometrical speaker identification, language identification, or linguistic studies about the features of human speech.

Traditionally, 2-dimensional feature extraction methods are used to derive features of each video frame which are then combined with a Hidden Markov Model (HMM) to obtain the most likely phoneme sequence. They suffer from a lack of features reflecting information of the entire time series due to the problem of not knowing the exact start positions and durations of each phoneme in advance. In contrast, this paper suggests a new method for extracting 3D features from video sequences which allows to recognise the spoken sounds better than comparable previous methods in the literature. 

The following Section~2 will explore relevant literature regarding  the topic and will outline the rationale for the proposed method based on the existing work. Sections~3 will then describe our approach in more detail, starting with the mouth area segmentation, followed by the feature extraction approach and the classification method. Section~4 details the conducted experimental work and Section~5 evaluates the results. Finally, Section~6 provides a conclusion by summarising the achievements of the paper and showing possible areas of applications and improvement. 

\section{Related Research}

Visual speech recognition requires knowledge and insight into the different areas contributing to successful implementations. Firstly, we need a basic understanding of human speech and its associated ambiguities and problems. Secondly, we will decide on the units of speech to be used for the experimental work. Thirdly, we review available mouth area segmentation techniques which pre-process the input aiming to find the lip shape or mouth area. Fourthly, we will examine possible extractable features which allow the differentiation between different units of speech. Finally, the section outlines applicable classifiers which can be trained and applied under the given circumstances to recognise an unknown speech sequence. In the course of presenting this related work, we 
outline the rationale for the proposed method.

\subsection{Human Speech Ambiguities}

Of the many different units of speech which can be produced by humans, only some are used in each language. Sets of indistinguishable units of speech are called allophones and together they constitute a phoneme, the minimal unit of speech that makes a perceived difference for speakers of a language \cite{Aronoff2003}. Moreover, the effects of co-articulation introduce another level of fuzziness. Co-articulation is the change of phones depending on the context. For example, in British English \textit{'that boy'} is often pronounced as \textipa{[Dap\,bAI]} instead of \textipa{[Dat\,bAI]}, changing the \textipa{/t/} for a \textipa{/p/} in anticipation of the following \textipa{/b/} (ibid.,~p.~176). One way to overcome this problem in recognition tasks is to use biphones or triphones (pairs or triplets of consecutive phonemes) which have a more distinct and unique pronunciation pattern \cite{Aleksic2004,Yu2008}. Yet another complication is that mouth movements and perceived sounds are not synchronised. Movements usually start before the actual sound and may end after the sound with an empirically measured average offset of about 40\,ms \cite{Sargin2007}.

Based on the described human speech ambiguities, we decided to counteract the audio-video time shift and the co-articulation effect by reverse-shifting the training data and by using biphones respectively. Both countermeasures can be easily incorporated, yet they have been neglected by many authors of visual speech recognition methods.

\subsection{Units of Speech}

Looking at the visual recognisable pose of the mouth, we can cluster all possible poses not distinguishable from another to visemes (similar to phonemes as clusters of indistinguishable units of speech). Different mappings from phonemes to visemes have been proposed based on linguistic analysis \cite{Jeffers1971,Pandzic2003}. However, there is no generally accepted mapping. In the context of visual speech recognition, some authors tried to create their own mappings based on automatic clustering of visual appearances of phonemes \cite{Neti2000,Hazen2004,Yau2007} or by clustering the appearances to elemental visemes without a clear mapping to phonemes \cite{Taylor2012}. Recent comparative studies showed that linguistic-based visemes have higher recognition rates on new test datasets \cite{Cappelletta2012}.

In order to prevent an unnecessary restriction of our method, we will train and test the system using English phonemes and the corresponding visemes of the linguistic-based Jeffers mapping \cite{Jeffers1971}.

\subsection{Mouth area segmentation}

Methods aiming at segmenting the lip shape, boundary or mouth area from the images of the input video can be divided in two main types: region-based and model-based approaches. Region-based approaches try to find the mouth area only and use a rectangular box or ellipse around the mouth as segmentation output. Model-based approaches try to fit a certain shape model of the mouth to the data which results in finding the outer and/or inner boundary of the lips.

Region-based methods can be further divided into three techniques: colour thresholding, colour clustering, and learned classifiers. Thresholding methods \cite{Hosseini2009,Cappelletta2010} use a colour channel of the picture to define thresholds over/under which pixels are considered as potential lip pixels. The largest connected area with potential lip pixels is then taken as the mouth. Clustering approaches also try to separate lip pixels from skin pixels based on colour, but use clustering techniques like k-means to aggregate the colours in the image to similar clusters. All pixels in the cluster with a mean colour closest to some given ideal lip colour are considered to be lip pixels \cite{Huang2010,Hassanat2010}. The third method is to manually label some regions in pictures as mouths in order to train a classifier that finds likely mouth regions in a picture automatically \cite{Duchnowski1995,Patterson2002,Castaneda2005}. All three approaches to region-based segmentation produce results of comparable quality and are able to find the correct mouth area in most cases. Reasons for failure often include too small colour differences between skin and lip areas or other areas of the image that resemble the lip colour more closely than the lips themselves. For some authors, this segmentation is only a preprocessing to a subsequent model-based segmentation allowing to restrict the search area to a small image patch \cite{Tian2000,Huang2010}.

Similarly, model-based approaches can be subdivided into three categories. Firstly: gradient-based approaches, secondly: active contour models (ACM) and thirdly: active shape and appearance models (ASM and AAM). Gradient-based approaches try to use image colour values and especially their gradients along horizontal and vertical lines to detect key points of the lips at maxima/minima or other statistically relevant positions of these values \cite{Kaucic1996,Mahdi2008,Li2009}. These key points are then used to fit polynomial functions between them as an approximated lip boundary. The second approach (ACM) utilises the gradient information as well by trying to improve the boundary iteratively from a starting position till it converges to the actual lip boundary. The iterative changes are driven by `external forces' which move the contour towards areas with high gradients and `internal forces' which try to maintain a smooth curve. This method was, e.g., used by \cite{Seyedarabi2006,Liu2010}. However, ACMs are very susceptible to the configuration parameters and the initial position. Hence, this method often converges to other boundaries which are not part of the lips. Some authors successfully combined the pure gradient-based estimations with ACMs achieving much better and more reliable results \cite{Eveno2004,Stillittano2013}. A third approach to model-based segmentation of lips is again based on a learning process \cite{Ong2008,Chitu2010,Newman2010}. Shape or appearance models of the whole face or the mouth area only (consisting of points and connections in-between) have been fitted to pictures manually. From this training data, few principal components of possible shapes or appearances can be extracted by Principal Component Analysis (PCA) which reduces the dimensionality of the search problem to a few parameters. Search techniques for global minimisation can then be applied to find the best model parameters for an image.

Independent of the applied segmentation method, the mouth position or lip boundary can be tracked in the consecutive video frames instead of being recalculated each time. A number of different tracking procedures have been used, including the Lukas-Kanade algorithm \cite{Tian2000,Eveno2004,Karlsson2012} which averages small changes in the proximity or probability-based tracking methods like Kalman Filters \cite{Kaucic1996} or Particle Filters \cite{Jiang2006} which assign probabilities to different movements. Some refined the tracking results with the methods used in the original segmentation process \cite{Eveno2004,Jiang2006,Seyedarabi2006}.

Since the proposed method entirely relies on pixel-based features, region-based mouth area segmentation should be sufficient for our case. However the susceptibility to small colour differences between skin and lip areas is reason to reject these approaches. On the other hand, model-based segmentation techniques take more information into account than just pixel colours. Therefore, they tend to produce more reliable matches than region-based segmentations. Consequently, our method is using a slimmed version of a model-based approach to find the correct mouth area, discarding any additional shape information which is not required afterwards. Thereby, we combine the robustness of the model-based approaches with the simplicity of the region-based output. Moreover, our approach applies a probabilistic tracking method which takes information of the entire video sequence into account and not just from the respective previous picture.

\subsection{Feature Extraction}

In order to distinguish different phonemes or words from another, effective features need to be extracted from the input data. The available methods can be divided in those which use one video frame at a time only and those which use the entire video working in the spatio-temporal feature space including information about the change over time.

The first type of features is usually obtained by reducing the dimensionality of the image data. One common method is the Principal Component Analysis (PCA) which projects the data on the most important eigenvectors in the feature space \cite{Chitu2007,Lan2010,Cappelletta2012}. A second method is to use the two-dimensional Discrete Cosine Transform (2D-DCT) which transforms the pixel data in a weighted sum of cosine functions of different wavelengths. The low frequency weights describe the most important information in the image and are therefore used as features \cite{Livescu2007,Abel2009,Pass2010}. The 2D-DCT is often combined with a subsequent PCA to find the most important frequencies automatically \cite{Saenko2009,Estellers2012}. Other applied dimensionality reduction methods are, for example, a Gabor Wavelet Transform converting the data in a weighted sum of Gabor functions \cite{Sujatha2010} or learned binary patterns which are selections of few pixels thresholded to binary values and taken as feature vector. Multiple different binary patterns were used to achieve an improved result \cite{Ong2011}.

Another possibility to extract features from the segmented mouth area is based on the shape and boundary of the lips.  These are usually reduced to a number of key points or contour points. From them, statistical features like coordinates, widths, heights, ratios, areas, or perimeters can be extracted \cite{Ibrahim2012,Kubanek2012}. Additionally, statistical features of the inner-mouth pixel data can be used like the area of dark pixels or the number of teeth or tongue pixels \cite{Hassanat2011,Singh2012}. Features of this category show equal performance on the subsequent classification process as pixel-based features. 

Most of the previously mentioned papers try to overcome the problem of not using temporal features by adding the first and second derivative of the extracted 2D-features in the video as additional feature sets. This adds information reflecting the change over time. However, this information still only shows aspects of  one specific point in time and not aspects of a whole time series. Real spatiotemporal features have been used rarely due to the problem of not knowing the  lengths of the speech units in continuous speech sequences. Phonemes can have very different lengths (even the same phoneme can be pronounced in different lengths) making it difficult to extract features from time series if the length of these series is not known a priori. Hence, these features have only been applied to cases in which words or phrases have been segmented already and only need to be recognised. Some authors use  a three-dimensional Discrete Cosine Transform (3D-DCT) to extract the most important frequencies in space and time \cite{Picard2010,Min2011}. Another approach is to use a 3D version of the already mentioned binary templates as a learned selection of pixels in 3D-space which reflect different properties of the time series. A histogram of the number of occurrences of different binary templates can be used as a feature vector \cite{Zhao2009}.

As far as known, nobody tried to use 3D features for general continuous recognition tasks based on phonemes yet, mainly due to the unknown speech unit lengths making it unclear which period of time to use for feature calculation. However, knowing that 3D features tend to perform better than 2D features on the same phrase recognition task \cite{Ong2011}, it is likely that 3D features will improve the overall accuracy on continuous recognition tasks as well. To overcome the problem of not knowing the sub-sequence positions and durations, our method calculates the features and the classification for each feasible sub-sequence. The 3D-DCT is our method of choice as it is easy to implement and the 2D-DCT is also one of the best methods to extract 2D features for continuous recognition tasks.

\subsection{Classifiers}

The traditional classifiers take a pre-specified number of features and output the most likely class the input belongs to. Due to the continuous nature of speech sequences with varying lengths, these methods can only be applied in cases in which whole sequences (isolated words or phrases) shall be recognised. Examples include the usage of Artificial Neural Networks \cite{Singh2012}, Support Vector Machines (SVM) \cite{Zhao2009}, Linear Discriminant Analysis \cite{Deypir2011}, k-Nearest-Neighbour \cite{Hassanat2011} and template matching \cite{Abel2009}. However, in most cases the authors aim to cope with continuous patterns whose durations are not known a priori. Therefore, they use probabilistic models which can track the probability of being in a specific state over time, allowing to extract the most likely sequence of states (e.g. phonemes) over a time series (like a video).  The most prominent example of this type of classifier is a Hidden Markov Model (HMM) which has been used by most authors  \cite{Lan2009,Chitu2010,Newman2010,Cappelletta2012}.

Our approach is exploiting the flexibility and easy prevention of over-fitting which is provided by a SVM. However, the classified sub-sequences need to be combined to find the most likely continuous phoneme sequence. This task is accomplished with a tailor-made HMM which models the possible transitions and uses the SVM classification output as observations.

\section{Methodology and Design}

\begin{figure*}

\centering
\includegraphics{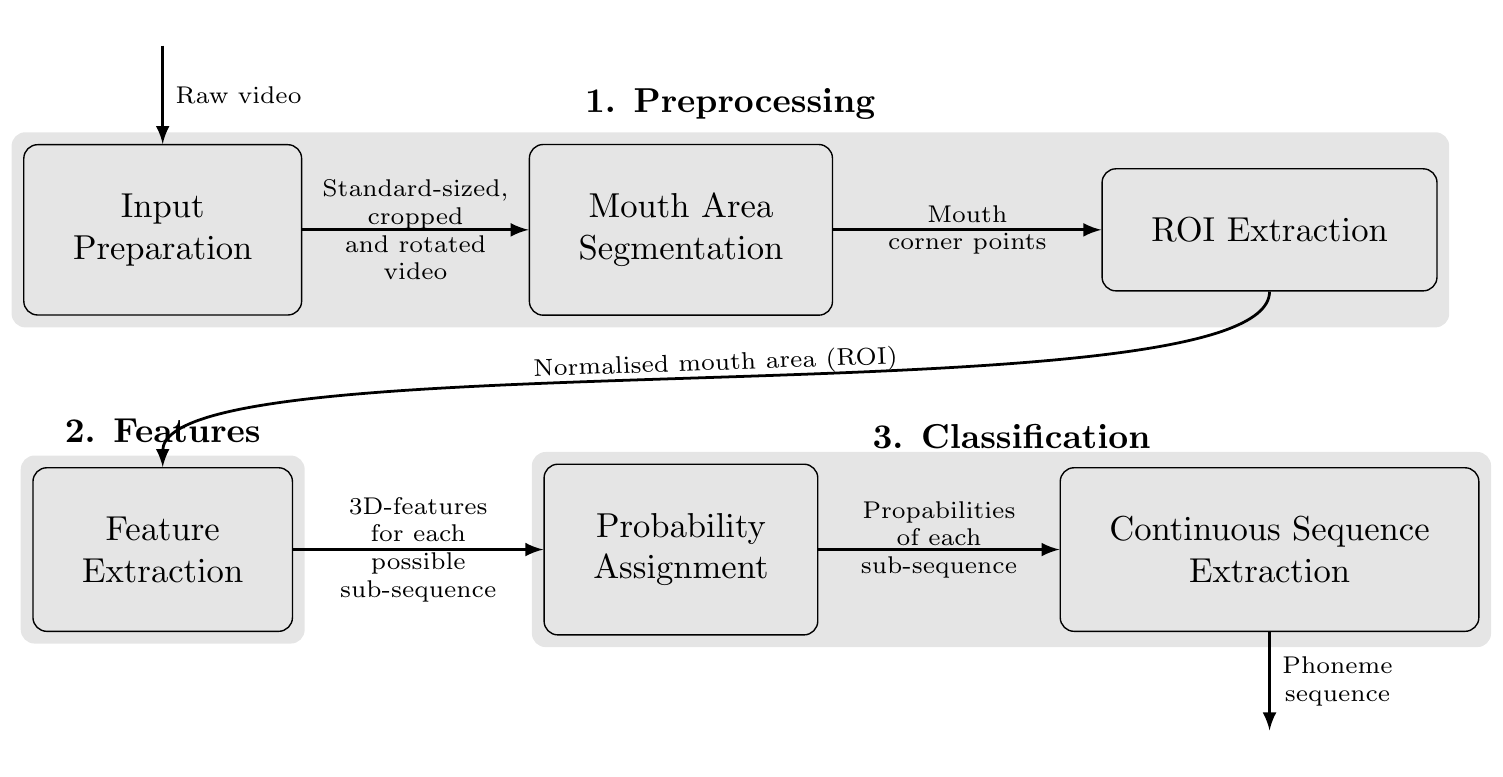}

\caption[Data flow of proposed method.]{Data flow of proposed method showing intermediate steps with their respective inputs and outputs.}
\label{fig_algoflow}
\end{figure*}

The algorithm can be divided into three major steps: Preprocessing, feature extraction and speech unit classification. The preprocessing includes the preparation of the video, the search for the mouth area in each image and the extraction of a Region of Interest (ROI) which comprises the normalised mouth area only. The second major step consists of the extraction of 3D features from the ROI for each possible sub-sequence. Finally, in the last major step, probabilities of being a specific phoneme are assigned to each sub-sequence and these probabilities are combined to calculate the most likely continuous series of phonemes. The algorithmic data flow is pictured in figure \ref{fig_algoflow} showing the major steps and their respective sub procedures with the input and output data of each step. Each of the major steps will be examined individually in the following subsections.

\subsection{Mouth Area Segmentation}

In the first stage of the algorithm, we aim to take the raw video input data and reduce that data to a ROI which only contains the required mouth area. Moreover, we want to ensure that the pre-processed output is consistent, does not contain outliers, and is normalised in such a way that the output is scale, location and rotation invariant to the input.

In summary, our proposed method consists of the following steps:
\begin{enumerate}
\item Input preparation by resizing all images to a standard size, finding the face symmetry line, and rotating and cropping the images based in this symmetry line.
\item Detection of the mouth area by applying a search technique that locates the inner lower lip on the symmetry line. Left and right mouth corner points are estimated based on the inner mouth gray-value gradient.
\item All mouth corner points are used to normalise the input to a ROI which is rotation, scale and location invariant.
\end{enumerate}
Each of these steps is completed for the entire video sequence before continuing with the next step. The following paragraphs will describe the design of each part of the algorithm in detail.

In the first part, the input data is the raw video consisting of 25 frames per second in which each frame is an image in the RGB colour space. To simplify the following lip and mouth detection, the images should have a standard size and they should be rotated and cropped such that the symmetry line of the face is exactly vertical and central in each image. This preparation helps as it limits the lip search area to the symmetry line and the gradient information necessary for the detection of the lips is most reliable if the face has a clear upright orientation.  

\begin{figure}
\centering
\includegraphics[width=\columnwidth]{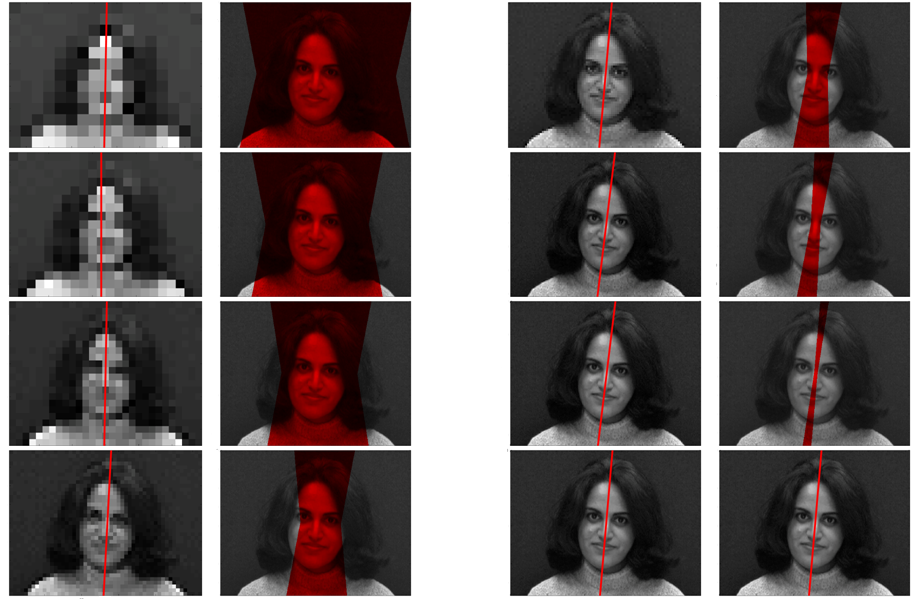} 

\caption[Illustrated example of the symmetry line extraction algorithm.]{Illustrated example of the symmetry line extraction algorithm. The figure shows 8 levels of the image pyramid. The left hand image in each pair shows the down-sampled image and the respective symmetry line that minimises the cost function in this image. The right hand image in each pair shows the remaining search area in the original image that corresponds to the line in the down-sampled image.}
\label{fig_symline}
\end{figure}

To find the symmetry line in the first frame of the video sequence, an image pyramid of the image is created in which each new image is 75\% the size of the previous one. The pyramid is computed till the width of the image is less than 20 pixels. Starting from the smallest image, possible symmetry lines are tested with a range of rotation angles and column indices. The quality of a symmetry line is evaluated by calculating the sum of squared differences of the pixel values on the left and the right side of the symmetry line. The best line which minimises the cost function is used as starting point in the next larger image. In each image of the pyramid the rotation angle is allowed to change by $\pm 1\degree$ and the symmetry line column by up to $\pm 2$ pixel. The cost function is always evaluated on 5 pixel columns on either side of the line. In the first image, the central column and a rotation angle of $0\degree$ is assumed. Using this recursive method, the symmetry line converges to its actual position in the original image without the need to test a large range of values for the angle and the position in the original image. Moreover, the method ensures that the symmetry condition is minimised on multiple scales preventing the line from converging on the eyes or between one eye and the nose as it would occasionally happen when the symmetry is computed in the original image only. For all subsequent images in the video, the position and angle of the symmetry line in the previous image is used as a starting point to find the new symmetry line directly in the full-size image. Again, the line is allowed to change by $\pm 1\degree$ and $\pm 2$ pixel from one frame to the next. An example of the process to find the symmetry line in the first image is shown in figure \ref{fig_symline}.

As further preparation, the image is cropped to $\pm 50$ pixel on either side of the symmetry line to reduce the amount of data. Moreover, a grey-level picture (or luminance picture) is calculated in which the minimum and maximum grey-value is mapped to 0 and 1 in order to enhance differences in the image. Additionally, the colour $U*lum$ is calculated for each image by multiplying $U$, which is the second colour in the CIE L*U*V* colour space, with $lum$, the value of the previously created luminance image. Both colours are needed for the next steps in the lip segmentation process.

\begin{figure}

\centering
\begin{minipage}[b]{0.4\columnwidth}
\centering
\includegraphics[height=\textwidth]{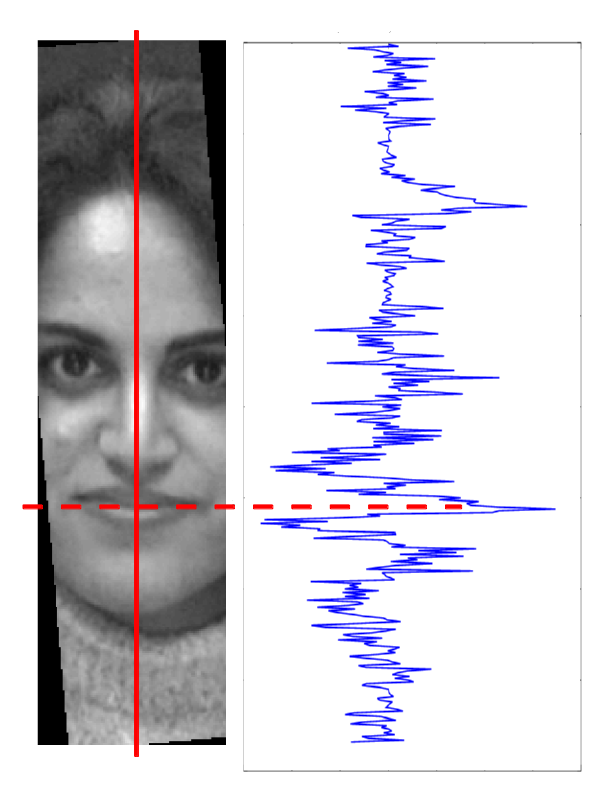}

(a) Inner lower lip detection
\end{minipage}
~
\begin{minipage}[b]{0.5\columnwidth}
\centering
\includegraphics[height=0.8\textwidth]{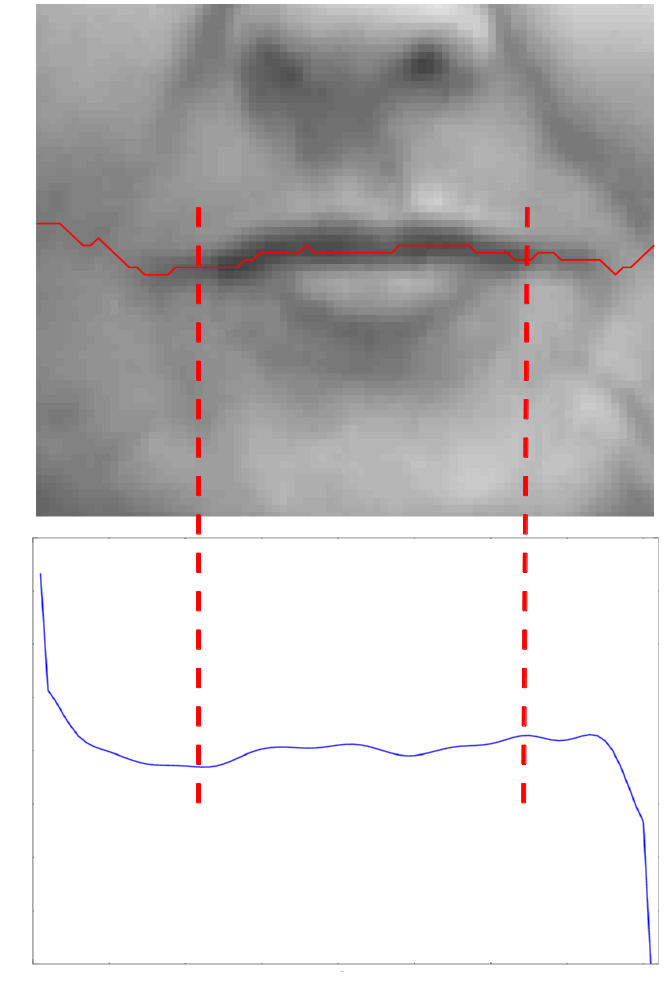}

(b) Mouth corner point detection
\end{minipage}

\caption[Exemplary graphs for the inner lower lip and mouth corner detection.]{Exemplary graphs for the inner lower lip and mouth corner detection. Figure (a) shows the cropped input with the symmetry line in red and the gradient graph of $U*lum$ along this symmetry line having a peak at the inner lower lip. Figure (b) shows the mouth area with the `darkest line' highlighted in red and the gradient of the smoothed luminance values along this line. The minimum gradient in the left part and the maximum gradient in the right part correspond to the mouth corner points.}
\label{fig_gradients}
\end{figure}

The second major step of the preprocessing is to find the lips and mouth corner points that allow the subsequent extraction of the ROI. In \cite{Stillittano2013}, the authors propose different colour gradients which highlight specific parts of the lips. Exploratory tests showed that the gradient of $U*lum$ for the inner lower lip is more reliable and more unambiguously than the gradients for other lip boundaries. As the full boundaries of the lips are not needed for the extraction of the ROI, our algorithm restricts its search to the inner lower lip using this gradient which should have its maximum at the lip boundary. To get a fast and easy estimation of possible intersections of the symmetry line with the inner lower lip boundary, we can compute the gradient in each image for the symmetry lines only. Figure \ref{fig_gradients}a shows that the gradient along the symmetry line peaks at this position in an example picture. 

Assuming equal priors for the initial position of the inner lower lip and having estimations of the likelihood for each position in the following images, we can apply a HMM to find the most likely sequence of positions at which the symmetry line and the inner lower lip boundary intersect. Each line index is a possible hidden state of the HMM. The observation probabilities are directly based on the gradient values in each image which are normalised to the range of $0$ to $1$. The transition probabilities try to model the assumption that the position of the intersection does not change significantly from one frame to the next. Therefore, the HMM uses the function values of a normal distribution function with a standard deviation of 8 pixels as transition probabilities. The Viterbi algorithm \cite{Viterbi1967} calculates the most likely sequence taking every included piece of information into account.

A similar approach is used to find the left and right mouth corner points from there. Based on the findings of \cite{Eveno2004} and \cite{Stillittano2013}, we know that the mouth corner points lay on the `minimal luminance line' between the upper and lower lip and the corners can be determined approximately by the gradient of the luminance values along this line. This `minimal luminance line' is not straight but follows the current shape of the mouth. It is constructed by starting with the darkest point just above the inner lower lip boundary on the symmetry line (in a range of -8 to +4 pixels from the inner lower lip) and extending the line to either side by adding pixels in the next column and $\pm 1$ row until the line has a total length of 81 pixels. Figure \ref{fig_gradients}b shows an example of a `minimal luminance line' together with the luminance gradient along this line highlighting the left and right mouth corner. Both authors in the mentioned papers claim that this gradient is not accurate enough to recognise the  transition from skin to inner mouth. However, using the same trick of applying a HMM to take all information from all video frames into account, we can reliably and consistently find the mouth corner points as well. The initial probabilities for each corner position are all 1 assuming equal priors and the transition probabilities are again based on a normal distribution function with a standard deviation of 2 pixels. The gradient values of the luminance colour normalised to the range of 0 to 1 are taken as observation probabilities. The combination of the gradient information and the restriction to small changes between images ensures a robust detection of the mouth corner points.

\begin{figure}
\centering
\includegraphics[width=\columnwidth]{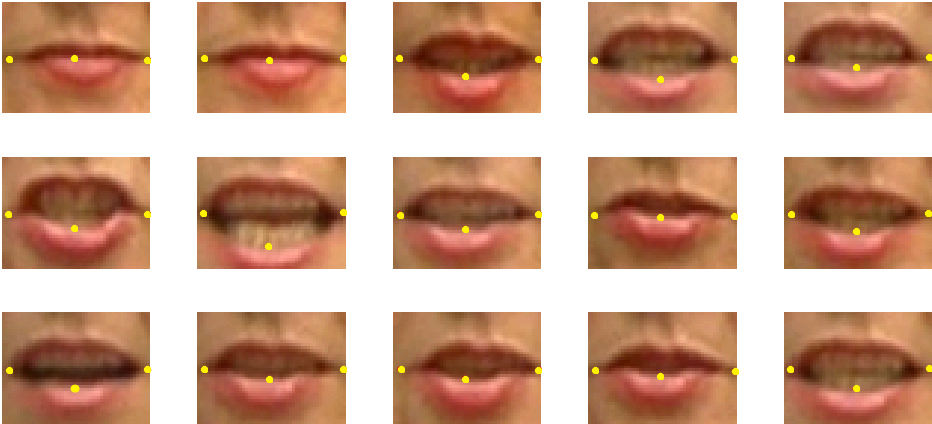}
\caption[ROI output of a video sequence.]{ROI output of a video sequence showing the first 15 frames of the normalised Region of Interest with yellow dots indicating the three keypoints used to find the ROI (left and right mouth corner, and intersection of inner lower lip with symmetry line).}
\label{fig_roi}
\end{figure}

The last major step of the preprocessing is to extract the ROI based on the mouth corners. The aim for the final output is a video sequence which only shows the mouth area, with the same image size for each frame and the condition that the video is aligned in such a way that only actual lip movements remain in the video and all other movements are not visible any more. This is achieved by rotating the image such that the connecting line between both moth corners is horizontal and central in the image. To resize the image, we will not force the corner point to constant positions in the video as this would eliminate all horizontal mouth width changes. Instead, we use the maximum mouth width (based on the corner points) to calculate a constant scaling factor for the entire sequence. By aligning the central points of the connecting line between the mouth corners we nevertheless ensure an output that only includes actual lip movements and no additional head movements under the assumption that the distance of the face to the camera roughly stays the same. Figure \ref{fig_roi} shows an example of the ROI output for a video sequence.

All in all, after cropping, rotating, and resizing the image we get a normalised input for the next step in which all interfering effects like face or mouth rotation, or changing positions have been removed. The process also includes a robust outlier prevention mechanism by applying HMMs. In the end, the result is rotation, position and illumination invariant to the input and also scale invariant in certain ranges that are still covered by the manually assigned parameters of the process.

\subsection{Spatiotemporal Feature Extraction}

The heart of the proposed method is its second major step which aims at extracting features which can be used for the subsequent classification of the uttered speech units. The feature extraction algorithm itself stays the same whether phoneme, visemes or other speech units are used, solely the resulting trained model of the classifier is different. Since the actual duration and start time of each speech unit is unknown, the feature extraction and classification is executed for each eligible sub-sequence. The deduction of the most likely sequence using the classification results will be part of the third and final step. The suggested method applies a 3D-DCT on each input sequence to extract a set of features based on the amplitudes of the most important frequencies in each direction. Another important property of our approach is that it tries to evade the major ambiguities of human speech, co-articulation and audio-visual time shift, by using time-shifted input features and by not only trying to recognise phonemes, but also pairs of consecutive phonemes (biphones) which are more unambiguously and distinct from one another.

\begin{figure*}
\centering
\includegraphics[width=\textwidth]{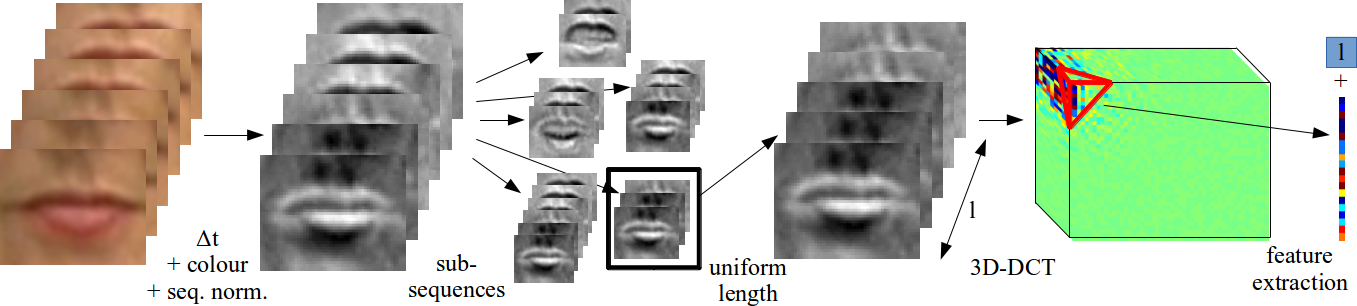}
\caption[Illustration of the feature extraction process.]{Illustration of the feature extraction process. The original video sequence is reduced to one colour, time shifted by $\Delta t$ and normalised with the sequence mean. Afterwards each eligible sub-sequence is extracted and each sub-sequence is normalised to a uniform length $l$. This input is used to calculate the 3D-DCT whose most important amplitudes extracted by a pyramid mask are taken as the feature vector. This vector is extended by the value $l$, the length of the respective input sequence.}
\label{fig_featureex}
\end{figure*}

Before conducting the 3D-DCT the input video sequence is reduced to one colour, is time-shifted by a time difference $\Delta t$ and the video sequence mean of each pixel in the frames is subtracted. The 3D-DCT works on one data value per pixel only, hence the reduction to one colour. There are various options for the colour including the Red, Blue or Green colour channel of the RGB input, the grey-value, or other colour combinations which have been used for lip detection purposes and are therefore likely to perform well in this context too. Moreover, the audio-video time shift which exists in videos of human speech can be counteracted by shifting the input in the opposite direction. Additionally, the between-sequence variance tends to be larger than the within-sequence variance which is the more important information as it resembles the actual lip movements. In order to diminish the effect of the between-sequence variance, we subtract the mean colour values of each pixel in the entire sequence. Hence, the remaining values only show positive or negative variations on the average mouth position. These three preparations are illustrated in the first step of figure~\ref{fig_featureex}.

Subsequently, all eligible sub-sequences are extracted. This is necessary since the classifier does not know the positions and durations of the speech units in advance and therefore needs to test all available options. The sub-sequences are obtained from each start position and for all durations between a minimal and maximal allowed duration. The minimum is always 40ms for phonemes or visemes, or 80ms for biphones, based on the minimum duration of phonemes in the training database. The maximum is a parameter that needs to be determined.

As further normalisation, each sub-sequence is interpolated to a constant length $l$. This ensures that the 3D-DCT features in time dimension are approximately equal for all samples of the same phoneme independent of the sample duration. However, this removes the information about the sub-sequence length which is an important clue since some phonemes tend to be spoken much faster than others. Hence, the sequence length will be added later as additional feature value. The extraction of the sub-sequences and the normalisation of each sub-sequence to a uniform length is illustrated in figure \ref{fig_featureex} as well.

After the mentioned normalisation and preparation steps, the subsequent 3D-DCT produces values whose main source of variation is the phoneme they are generated from. Similar to the triangular mask which is often used to extract the most important frequency amplitudes from 2D-DCT images \cite{Cappelletta2012}, a pyramid mask is used to obtain the amplitudes of the low frequency cosine waves in all directions which contain the essence of the video sequence. 

The resulting feature vector is extended by the length $l$ of the sub-sequence to include this knowledge as it was previously removed from the data. The 3D-DCT output, the pyramid extraction mask and the final feature vector are shown in figure \ref{fig_featureex} too. 

The whole process depends on a set of parameters which need to be determined with experimental tests. These parameters are the audio-visual time shift, the colour to be used for feature extraction, the length of this uniform duration and the size of the pyramid feature extraction mask. 

\subsection{Speech Unit Classification}

The remaining task of our proposed method is to classify the features of each sub-sequence and use the probabilities given by the classifier to determine the most likely continuous sequence of phonemes or visemes.

\begin{figure}
\centering
\includegraphics[width=\columnwidth]{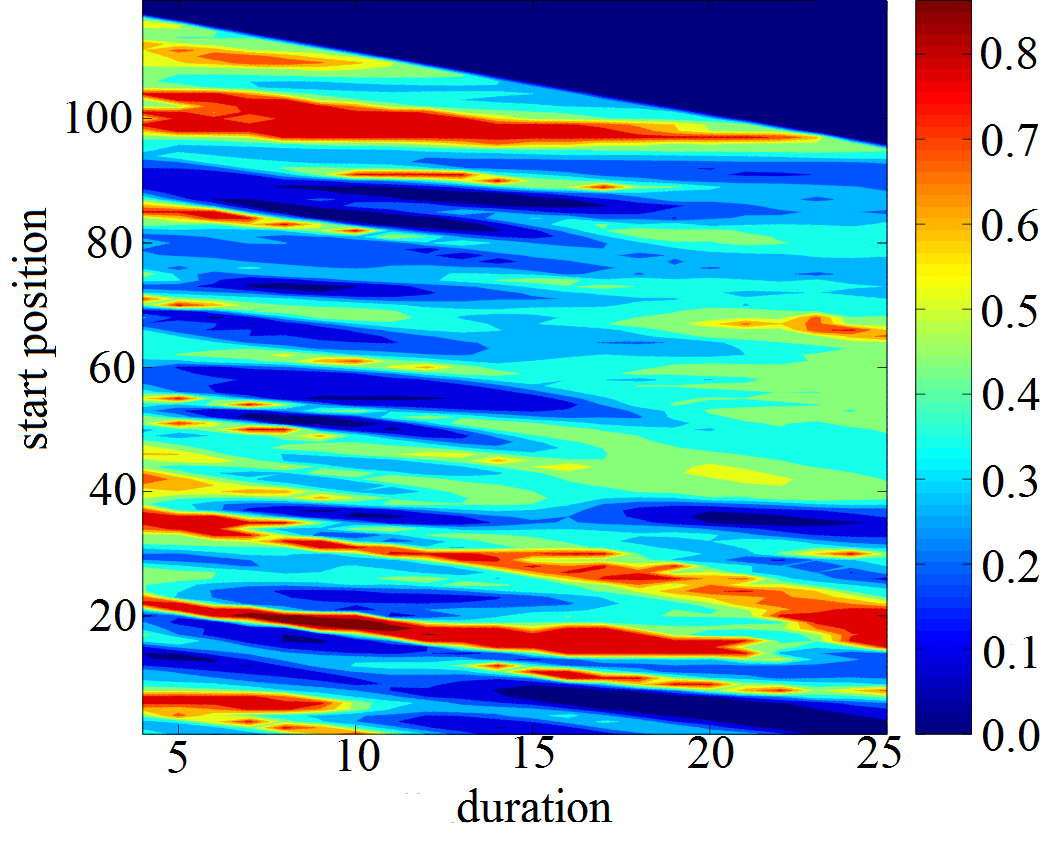}
\caption[Classification output illustrated as probability grid.]{Classification output illustrated as probability grid of an example sequence. The probability for each start position and duration pair is colour-coded as indicated in the colour-bar. The figure shows the probability grid for the phoneme \textipa{[\ae]}.}
\label{fig_svmprobs}
\end{figure}

The sub-sequence classification is done with a multi-class SVM. Before these feature vectors can be classified by the SVMs, they need be standardised again to ensure all values fall approximately within the range of $-1$ to $1$ which is a necessary precondition for a SVM to work well. All feature vectors are therefore subtracted by the mean and divided by the standard deviation of all feature vectors obtained from the entire training data set. These mean and standard deviation values are part of the trained model and are reused for each new unseen feature vector. SVMs are trained for each required output class (phonemes, visemes, biphones, etc.) using all features vectors of all sequences of the respective class in the training data set as input for class $1$ and all feature vectors of all other classes in the training data set as input for class $-1$. A separate cross-validation set needs to be applied to test different parameters preventing over-fitting by choosing the parameter set with the best performance on the cross-validation data set. Each SVM is furthermore trained to output probabilities for the input of being the respective class $1$. Thus, the output for each phoneme (or any other unit of speech) can be represented as in figure \ref{fig_svmprobs} with a grid of probabilities for this phoneme for each possible start position and duration. 

The classification process can be trained and executed for different sets of classes like phonemes, biphones, visemes, or viseme pairs (bi-visemes). The resulting probabilities will be combined in the final step of the algorithm, the continuous sequence extraction.

This task is accomplished with a tailor-made HMM which models the possible transitions and uses the classification output as observations. Alternative approaches for this concrete task are not available as this is the first method to apply 3D features and the associated probability output to continuous visual speech recognition tasks. Nevertheless, there are general purpose approaches using special variations of HMMs which can handle state durations as summarised in \cite{Johnson2005}. However, these methods try to deal with normally distributed probabilities of durations and are therefore more general than actually needed. As we already have exact information about the probabilities for each duration, we have more data than can be easily incorporated in these models. Therefore, a specially designed standard-type HMM is sufficient enough to obtain the desired result.

The remaining question is how to design the HMM such that it models all desired effects correctly. At each point in time, we get $N \times D$ new observations, in which $N$ is the number of classes (e.g. phonemes) and $D$ is number of possible durations of a sub-sequence. Each observation is the probability that the sequence of a certain duration and a certain class starts at the current point in time. In order to incorporate all cases without combining multiple cases and observations, we need at least the same number of hidden states in the HMM. Consequently, the HMM has at least $N \times D$ states $c_d^t$ each representing a sequence starting at the point in time $t$ of duration $d$ and class $c$. However, using these deliberations alone, it remains unclear how to model the transitions between the states. 

\begin{figure*}

\centering
\begin{minipage}[b]{0.19\textwidth}
	\centering
	\includegraphics{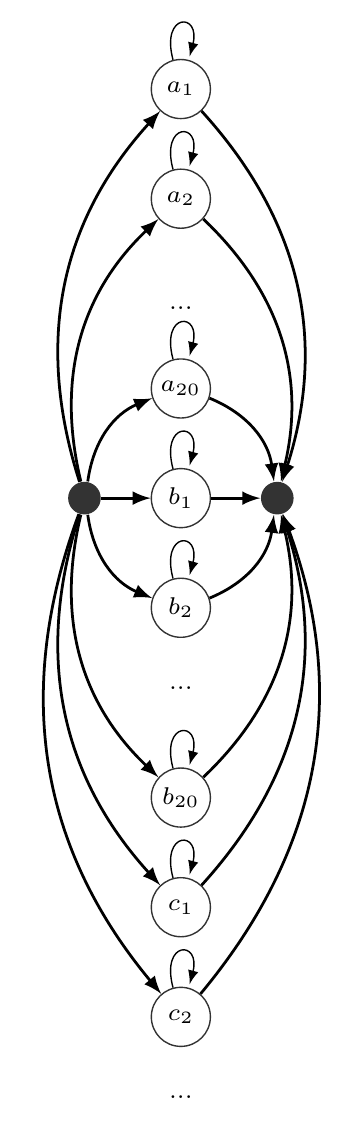}
	
	(a)
\end{minipage}
\begin{minipage}[b]{0.15\textwidth}
	\centering
	\includegraphics{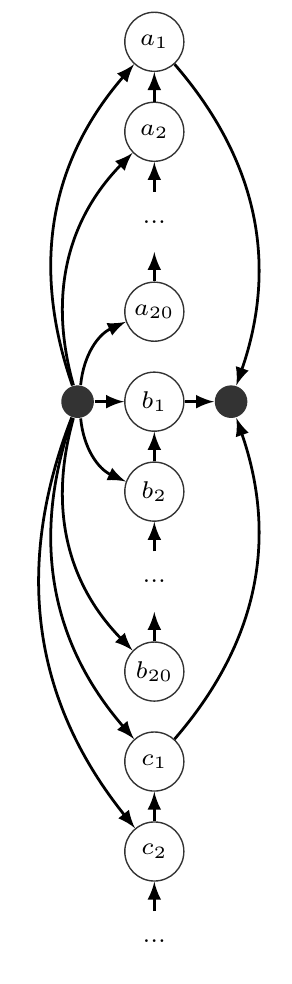}
	
	(b)
\end{minipage}
\begin{minipage}[b]{0.43\textwidth}
	\centering
	\includegraphics{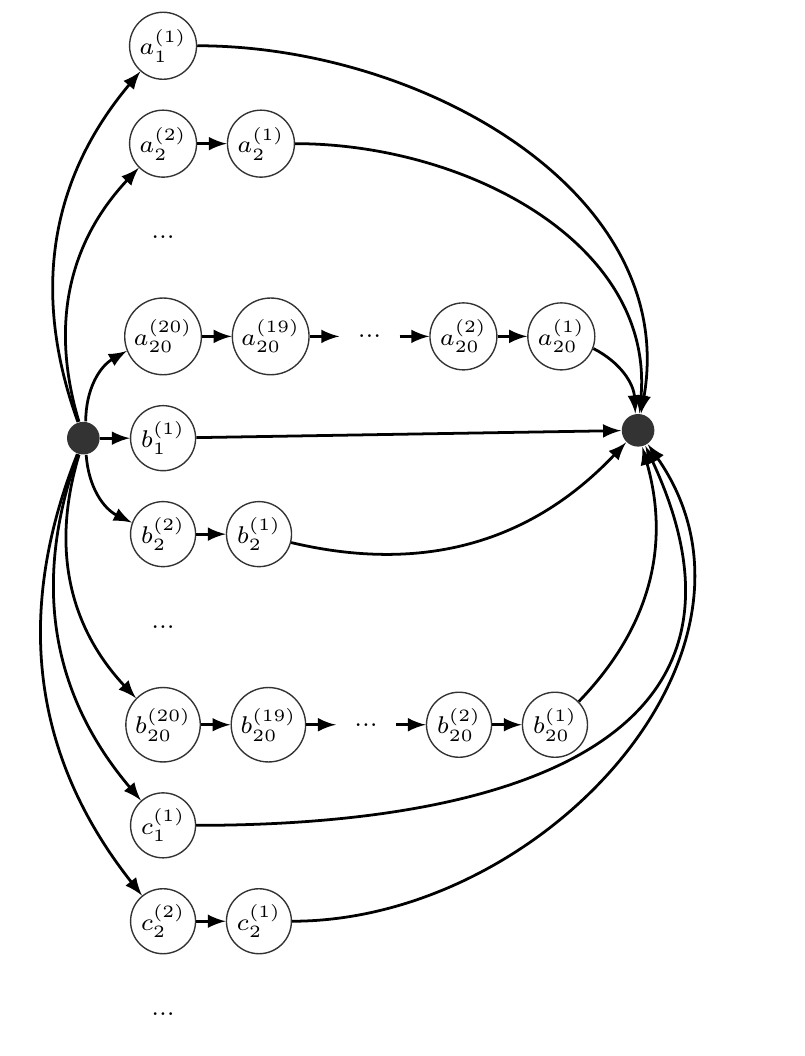}
	
	(c)
\end{minipage}
\begin{minipage}[b]{0.2\textwidth}
	\centering
	\includegraphics{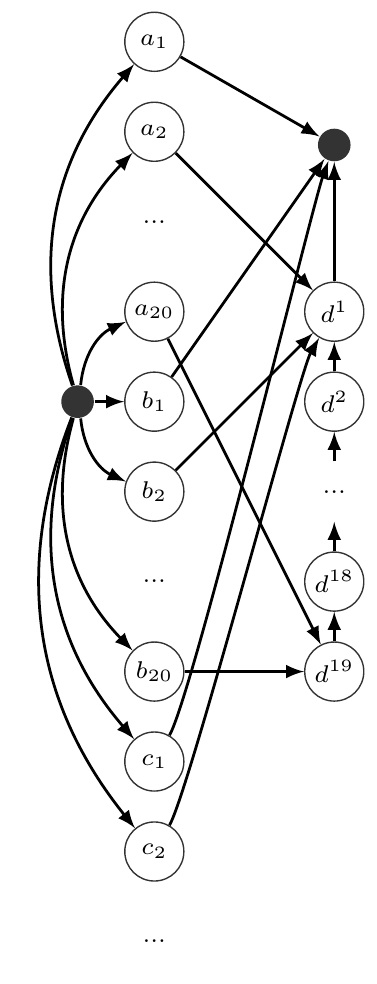}	
	
	(d)
\end{minipage}

\caption[Options for HMM transitions to model continuous visual speech recognition.]{Options for HMM transition configurations to model the continuous visual speech recognition. Option (a) shows a transition configuration using looping transitions to stay in the same state. Option (b) uses transitions to the states of the same class and decreased duration. Option (c) introduced additional dummy states for each class-duration pair. Option (d) uses common dummy states used by all other states together. Only options (c) and (d) fulfil all requirements of the task. The hidden states are marked with the corresponding class $a$, $b$, or $c$ and the associated duration in the subscript. The maximum duration is $20$. Dummy states are marked equally or with $d$ using additional superscripts to indicate the index. The black dots are used as start and end points before and after which the presented transition process continues anew. }
\label{fig_hmmmodels}
\end{figure*}

Figures \ref{fig_hmmmodels}a and \ref{fig_hmmmodels}b show two options of which transitions to allow from one point in time to the next (all transitions with a transition probability greater than zero). Option~(a) uses loops which allow to stay in the same state during the duration of the associated phoneme. However, this can be rejected due to multiple reasons. As the transition probabilities of staying in the state or leaving it are constant and only based on the previous state, it is not possible to force the HMM to stay an exact amount of time in one state. Moreover, each state combination of class and duration can start at any point in time making it impossible to model all simultaneously. For example, a phoneme \textipa{[\ae]} with a duration of 15 frames might start at time $t=5$ or at $t=7$ having different probabilities for either case. That makes it unclear which of the probabilities to use as observation probability in the following time steps. Option~(b) forces the phonemes to the specified duration by only allowing transitions to the next state of the same phoneme and a decreased duration. However, the transition model still suffers from the problem of not knowing which probability to use as observation probability since the original probability stays valid for the entire duration of a phoneme.

Figures \ref{fig_hmmmodels}c and \ref{fig_hmmmodels}d show two further options which both overcome the mentioned problems. However, both use additional hidden states. Option~(c) uses dummy states to model the duration of each case separately, whereas option~(d) uses common dummy states for all cases together. The aspired duration of the phonemes is enforced in both options by having only one possible transition from each state obliging the HMM to comply with the duration. For option~(c), all hidden states can have the respective observation probability for a phoneme of the specified class and specified duration having started at the given point in the past. Thereby, all observations can be modelled correctly and the transition model is therefore sufficient to legitimately find the most likely sequence of phonemes. However, the option uses many hidden states in the order of $\mathcal{O}(ND^2)$. On the contrary, option~(d) uses only marginally more states than minimally necessary. However, it is not possible to include all probabilities in the way as it would have been done in option~(c). The common dummy states are used for all classes and are therefore not able to model any specific probability. Knowing that the observation probabilities are all multiplied together in the course of calculating the Viterbi algorithm \cite{Viterbi1967}, we can apply the trick of using the total product of all probabilities of one phoneme of a class and duration as observation probability for the first hidden state. If the observed probability of starting a phoneme $c$ of duration $d$ at time $t$ is $P(o|c_d^t)$, the new observation probability is $P(o|c_d^t)^d$, since this probability would have been multiplied by itself $d$ times. The observation probabilities of the dummy states can then be equal to $1$ without loosing any information. Thus, based on the algorithmic knowledge of the Viterbi algorithm \cite{Viterbi1967}, we are able to model all observations correctly while using as few hidden states as possible.

Assigning equal priors to the initial states and transition probabilities of $1$ to all indicated transitions in figure \ref{fig_hmmmodels}d and $0$ otherwise, we have made all necessary preparations to run the Viterbi algorithm \cite{Viterbi1967} to find the most likely sequence of states. The most likely sequence of states can then be squeezed to the most likely sequence of phonemes by removing all dummy states. A further advantage of this method is that it can be used in parallel for phonemes and biphones combining the possibilities of finding single phonemes or pairs. It can even be adjusted to allow different durations for each class as transitions to the relevant dummy states can be added or removed as necessary.

In summary, our proposed method uses a slimmed version of a model-based lip segmentation to find a Region of Interest in each image. These ROIs are normalised and used as input to a 3D-DCT feature extractor which in turn produces the input for SVMs to classify between different classes. The probabilities for each phoneme, biphone (or viseme and bi-viseme respectively) can be combined to find the most likely overall sequence using a tailor-made HMM. The method introduces multiple improvements to other existing approaches including the enhanced lip and mouth tracking with HMMs using all available information at once, the usage of 3D information for continuous sequence recognition by taking all possible start positions and durations and combining them with a tailor-made HMM, and also by taking care of the special ambiguities of human speech, audio-visual time shift and co-articulation, which have been neglected by many other reported methods.

\section{Experimental Work}

The previous sections investigated the literature and specified the details of the proposed method. The current section describes the experimental work done to implement and to evaluate our approach. The first part shows the considerations and necessary steps for the input preparation, including the database set-up and the training set selection. The subsequent three subsections will describe the experimental findings of the mouth area segmentation, the classification, and the continuous sequence extraction experiments. This includes the description of parameter optimisations and further intermediate as well as final results on the test sets.

\subsection{Input Preparation}
\label{chap_inputprep}

This subsection will describe the necessary input preparations made before implementing the proposed method. At first, we will look at the decision for an audio-visual database. Secondly, the division of this database into training and test sets is portrayed.

The decision for a training and test database is driven by the need to have examples of all phonemes of the English language in a natural, unconstrained setting which allows to test the method for general purpose applications. TIMIT sentences \cite{Zue1990} are the most promising input type as they feature all variations of phonemes equally and are therefore a well-balanced training input for a phoneme classifier. However, the large database AV-TIMIT \cite{Hazen2004} is legally restricted to uses within the Massachusetts Institute of Technology. Hence, the smaller VidTimit database \cite{Sanderson2009} is used which has, however, no timed transcriptions of phonemes. Yet, the audio signal can be segmented reliably into phonemes using the force-alignment mode of a trained speech recognition HMM. The process is outlined in \cite{Gorman2007}. Segmenting the audio data into phonemes to use the results on the video data is a common process which was also used by others \cite{Abel2009,Cappelletta2012}. The correct segmentation into phonemes has been verified by manual random sample examination.

\begin{table}
\resizebox{\columnwidth}{!}{
\centering
\begin{tabular}{cc}
\toprule
ID & Sentence text\\
\midrule
sa1 & She had your dark suit in greasy wash water all year. \\
sa2 & Don't ask me to carry an oily rag like that. \\
si1542 & The feet wore army shoes in obvious disrepair. \\
si2172 & Strength began to zip up and down his chest. \\
si912 & You may amaze yourself and acquire a real knack for it. \\
sx102 & Special task forces rescue hostages from kidnappers. \\
sx12 & Will Robin wear a yellow lily? \\
sx192 & Straw hats are out of fashion this year. \\
sx282 & The tooth fairy forgot to come when Roger's tooth fell out. \\
sx372 & That diagram makes sense only after much study.\\
\bottomrule
\end{tabular}
}
\caption[Examples for TIMIT sentences used in the VidTimit database.]{Examples for TIMIT sentences used in the VidTimit database. The first two sentences, sa1 and sa2, are recited by each person.}
\label{tab_timitsen}
\end{table}

The VidTimit database consists of video sequences of 43 Australian subjects (19 female and 24 male) each reciting 10 short TIMIT sentences. The first two sentences are always the same, the other eight sentences are different for each person. Table \ref{tab_timitsen} shows some examples of the recorded sentences. The recordings of each person were made in three sessions to ensure different clothes, make-ups and moods. Additionally, the camera distance was varied between recordings. The scene was well-illuminated and recorded with a video camera at 25~fps. The results, however, are stored frame-wise as JPEG images each 384~x~512 pixels in size.

In order to train and test the algorithm on separate data, we divided the database into a training and a test set. In line with previous work \cite{Cappelletta2012,Ramage2011}, $k$-fold testing has not been used due to the large amount of data and the associated long training times. Therefore, eight persons of the VidTimit database are chosen arbitrarily for testing (18.6\% of the database). The remaining 35 persons are used for training and optimisation. The selected person IDs for the test set are \textit{fcmr0}, \textit{fcrh0}, \textit{fdac1}, \textit{fdms0}, \textit{fdrd1}, \textit{fedw0}, \textit{felc0} and \textit{fgjd0}. Most other authors defined their test set as an arbitrary collection of video sequences across all available persons \cite{Cappelletta2012,Ramage2011,Newman2010}. We consider our approach more realistic as the persons eventually used for testing have never been used for training and optimisation (not even other videos of the same person).

The data sets of the remaining 35 persons are used to train and to optimise the algorithm with single speech units as well as entire videos. Since these tasks are independent of each other, there is no need to further divide the training set for this reason. However, the total set of single phoneme samples is further divided into a training and a cross-validation set used to optimise parameters of the algorithm. Therefore, 20\% of the samples of each phoneme are taken as cross-validation data and the remaining 80\% as the actual training data samples.  Table \ref{tab_datacount} shows statistics about the number of available training samples for each phoneme. It also shows the transcription to English characters used for each phoneme instead of the official IPA (International Phonetic Alphabet) character. The transcription follows the system of the ARPAbet \cite{Shoup1980} to represent phonemes of the English language.

\begin{table*}
\resizebox{\textwidth}{!}{
\centering
\begin{tabular}{ccccc|ccccc|cccccc}
\toprule
IPA & ARPAbet & \# train & \# cv & \# test & IPA & ARPAbet & \# train & \# cv & \# test & IPA & ARPAbet & \# train & \# cv & \# test \\
\midrule
\textipa{S} & SH & 128 & 28 & 38 & 
\textipa{i} & IY & 454 & 102 & 136 &
\textipa{h} & HH & 139 & 31 & 38 \\ 
\textipa{\ae} & AE & 371 & 83 & 92 &
\textipa{d} & D & 369 & 88 & 96 & 
\textipa{j} & Y & 137 & 33 & 39\\
\textipa{O} & AO & 230 & 53 & 56 & 
\textipa{r} & R & 541 & 133 & 135 &
\textipa{A} & AA & 235 & 51 & 61 \\ 
\textipa{k} & K & 377 & 83 & 103 &
\textipa{s} & S & 453 & 99 & 130  & 
\textipa{u} & UW & 186 & 43 & 55\\
\textipa{t} & T & 590 & 130 & 171 & 
\textipa{I} & IH & 452 & 102 & 129 &
\textipa{n} & N & 523 & 119 & 141 \\ 
\textipa{g} & G & 153 & 37 & 40 &
\textipa{w} & W & 201 & 49 & 66 & 
\textipa{3} & ER & 269 & 60 & 60\\
\textipa{l} & L & 455 & 103 & 128 & 
\textipa{oU} & OW & 126 & 30 & 41 &
\textipa{m} & M & 272 & 62 & 76 \\ 
\textipa{OI} & OY & 40 & 11 & 12 &
\textipa{AI} & AY & 156 & 35 & 39 & 
\textipa{D} & DH & 195 & 44 & 63\\
\textipa{b} & B & 155 & 38 & 41 & 
\textipa{2} & AH & 694 & 173 & 224 &
\textipa{v} & V & 131 & 28 & 40 \\ 
\textipa{f} & F & 162 & 36 & 40 &
\textipa{z} & Z & 250 & 58 & 62 & 
\textipa{T} & TH & 37 & 9 & 9\\
\textipa{p} & P & 188 & 41 & 57 & 
\textipa{E} & EH & 212 & 50 & 50 &
\textipa{eI} & EY & 114 & 28 & 38 \\ 
\textipa{N} & NG & 62 & 14 & 21 &
\textipa{tS} & CH & 41 & 9 & 10 & 
\textipa{U} & UH & 32 & 7 & 9\\
\textipa{Z} & ZH & 17 & 3 & 5 & 
\textipa{dZ} & JH & 41 & 9 & 16 &
\textipa{aU} & AW & 49 & 11 & 7 \\ 
\textipa{\o} & sil & 1123 & 266 & 375 & 
& & & & & 
& & & & \\
\bottomrule
\end{tabular}
}
\caption[Counts of each phoneme in the training, cross-validation and test data sets.]{Counts for each phoneme in the data set. The phonemes are described by their official IPA symbol and the ARPAbet transcription used throughout the implementation. The counts for the training data set (\# train), the cross-validation set (\# cv) and the test set (\# test) are given.}
\label{tab_datacount}
\end{table*}

\subsection{Mouth Area Segmentation Experiments}

\begin{figure}
\centering
\includegraphics[width=.9\columnwidth]{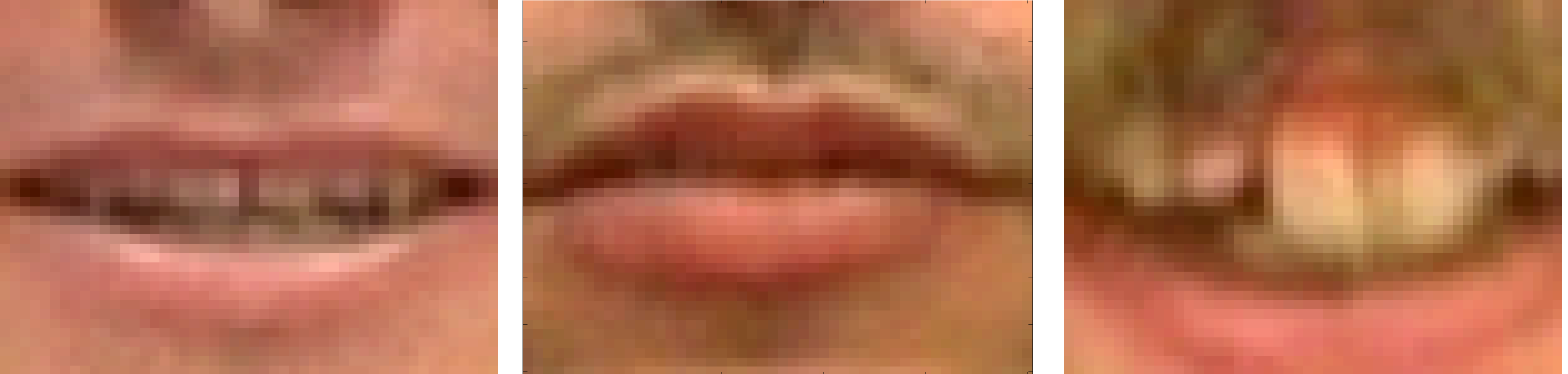}
\caption[Sample results of the mouth area segmentation algorithm.]{Sample results of the final mouth area segmentation algorithm.}
\label{fig_segmentfinal}
\end{figure}

The first part of the experimental work focuses on the preprocessing and mouth area segmentation. All parameters as they have been mentioned in the method description were found by trial-and-error tests on the training data set. An automatic optimisation for this part of the algorithm was not possible as no ground truth of the lip position is given in the database. Figure \ref{fig_segmentfinal} illustrates the results achieved with the proposed method. Note that even with the expansive moustache of the right person in the figure the segmentation produced reliable results.

In total, our lip segmentation approach segmented 404 of the 430 video sequences in the VidTimit database in an acceptable way. This has been checked manually to ensure the usefulness of the input for the next steps. The remaining 26 videos with segmentation errors have been adjusted by forcing the HMM for the inner lower lip detection to a specific line in the first frame, which is sufficient enough for the rest of the video to be correct as well.

\subsection{Classification Experiments}

The process of extracting features and training the SVM classifiers is the main part of the algorithm. In this procedure the main problem is the adjustment and optimisation of all parameters which influence the result. In principle, this obstacle is overcome by testing a range of possible values for each parameter on the training data set and checking the performance on the cross-validation set. The performance of the probability output for each phoneme or viseme is measured in the proportion of correctly classified cross-validation samples, where `correctly classified' means that the sample got its highest probability for the correct phoneme or viseme respectively. For most parameters the optimisation was done for visemes, since the calculation is usually faster as there are fewer visemes than phonemes. Because all of the parameters represent properties of the input data, it is unimportant whether they are optimised based on the phoneme or viseme accuracy. Therefore, the optimised parameters can be used for either case afterwards.

The visemes used in the experiments were based on the Jeffers mapping \cite{Jeffers1971} shown in table \ref{tab_visemes}, a mapping which was found to have the highest recognition rates in visual speech recognition \cite{Cappelletta2012}. The unknown parameters, which have been optimised in the described way, are:
\begin{enumerate}
\item $\Delta t$, the audio-visual time shift by which the ROI is shifted before the features are extracted.
\item The colour of the ROI used for feature extraction. Available options are: Red, Green, Blue, Grey-value, $U$ (of CIE L*U*V*), $U*lum$, and Pseudo-Hue (proposed colour for lip segmentation in \cite{Eveno2004}).
\item The uniform input length $l$ to which all sub-sequences are normalised. 
\item The size of the feature extraction pyramid. The pyramid size is determined by the number  $s$ of values in each dimension. The actual number of features then is $\frac{1}{6}s(s+1)(s+2)$ based on the volume formula of a square pyramid.
\item The SVM parameters $C$ and $\sigma$. Both parameters are not problem-specific and are therefore determined by cross-validation in each training process anew. The utilised SVM calculation software uses a parameter $\gamma$ instead of $\sigma$ which can be calculated by $\gamma=\frac{1}{2\sigma}$.
\end{enumerate}
\begin{table}
\centering
\begin{tabular}{p{1cm}p{2.2cm}|p{1cm}p{2.2cm}}
\toprule
Viseme & Phonemes & Viseme & Phonemes \\
\midrule
/A & F,V & /G & OY, AO \\
/B & OW, R, W, UH, UW, ER & /H & S, Z  \\
/C & B, P, M & /I & AA, AE, AH, AY, EH, EY, IH, IY, Y \\
/D & AW & /J & D, L, N, T \\
/E & DH, TH & /K & G, K, NG \\
/F & CH, JH, SH, ZH & /L & sil\\
\bottomrule
\end{tabular}
\caption[Jeffers viseme mapping used to map phonemes to visemes.]{Jeffers viseme mapping used to map phonemes to visemes. The visemes /A to /L correspond to the indicated phonemes in the second column. The phoneme HH is not assigned to a viseme.}
\label{tab_visemes}
\end{table}

\begin{figure}
\centering

\begin{minipage}[b]{\columnwidth}
	\centering
	\includegraphics{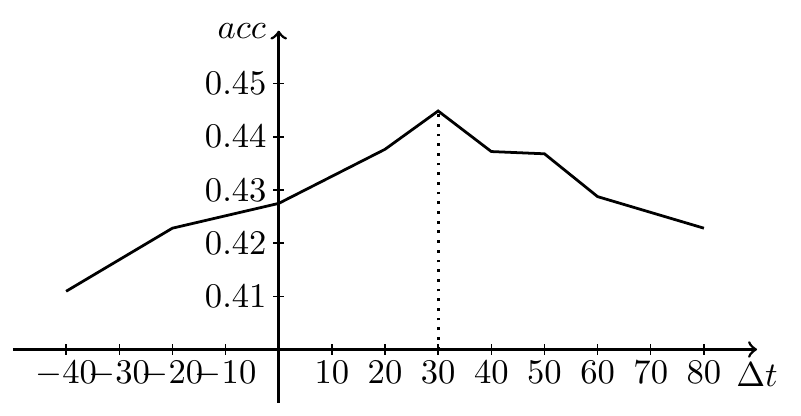}
	
	(a)
\end{minipage}

\begin{minipage}[b]{0.5\columnwidth}
	\centering
	\includegraphics{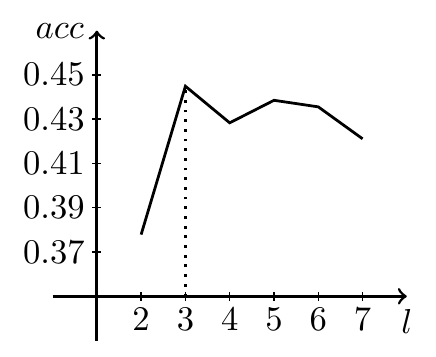}
	
	(b)
\end{minipage}
\begin{minipage}[b]{0.45\columnwidth}
	\centering
	\includegraphics{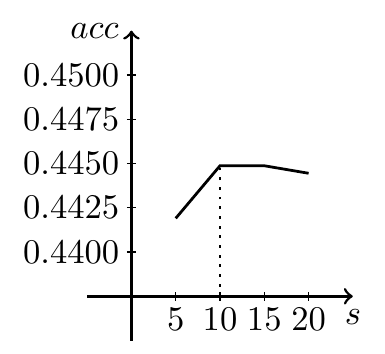}
	
	(c)
\end{minipage}

\resizebox{\columnwidth}{!}{
	\centering
	\includegraphics{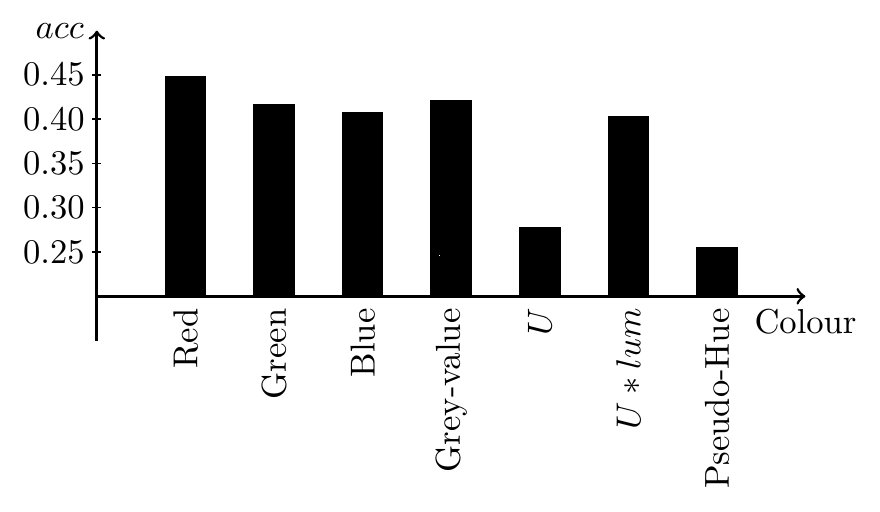}
}	
	(d)

\caption[Cross-validation results for various feature extraction parameters.]{Cross-validation results for various feature extraction parameters: audio-visual time shift $\Delta t$ in ms~(a), uniform input length~$l$~(b), the size of the feature extraction mask~$s$~(c) and input colour~(d). The accuracy values $acc$ show the proportion of correctly classified visemes in the cross-validation data set.}
\label{fig_crossvalall}
\end{figure}

Figure \ref{fig_crossvalall} shows the result of the cross-validation process in graphs depicting the cross-validation performance when the system is trained with different parameters. Only one parameter is varied in each case (plus the SVM parameters). All other parameters are held constant at their best value. As a result, it was found that the best $\Delta t$ is 30~ms, which approximately corresponds to the findings of \cite{Sargin2007} who specifically examined time shifts and found the highest correlation at a time shift of 40~ms. The colour with the best performance is Red, with the interesting outcome that colours specifically designed to detect lips by high gradient values ($U*lum$ or Pseudo-Hue) do perform worse when used for the feature extraction of the lips. The performance in terms of the uniform sequence length stays roughly constant for lengths of 10 frames or more. To reduce the size of the required memory, 10 frames are chosen as the best parameter. Finally, the size of the feature extraction pyramid has its peak performance at $s=3$ with slightly decreasing performance values for larger feature sets. This is due to the fact that higher-dimensional feature spaces are not as easily separated into the output classes as lower-dimensional feature spaces making the performance worse although more information is used. Therefore, the final feature vectors have 11 features ($\frac{1}{6}\cdot 3\cdot 4\cdot 5+1$) with one feature added for the length of the sequence. The cross-validation result for the SVM parameters has its performance peak at the parameter value pair of $C=2^6=64$ and $\sigma=\frac{1}{2\gamma}=\frac{1}{2\cdot 2^{-7}}=64$.
All these optimised parameters together are then used to train the classifier on the training data set for the probability output for phonemes, visemes, biphones, and bi-visemes.

The source code to calculate the 3D-DCT is based on \cite{Garcia2009}. The SVM training and classification, including the probability calculation, was done with the LibSVM package for Octave \cite{Chang2011}. 

\subsection{Continuous Sequence Recognition Experiments}

The final part of the experiments comprises all tests with the continuous sequence recognition. The tailor-made HMM has yet another parameter which needs to be optimised: the maximum duration of a phoneme or a biphone. Again, cross-validation is used to find the optimal value by checking the performance of different duration limits on the full sequence cross-validation set.

\begin{figure}

\begin{minipage}[b]{\columnwidth}
	\centering
	\includegraphics{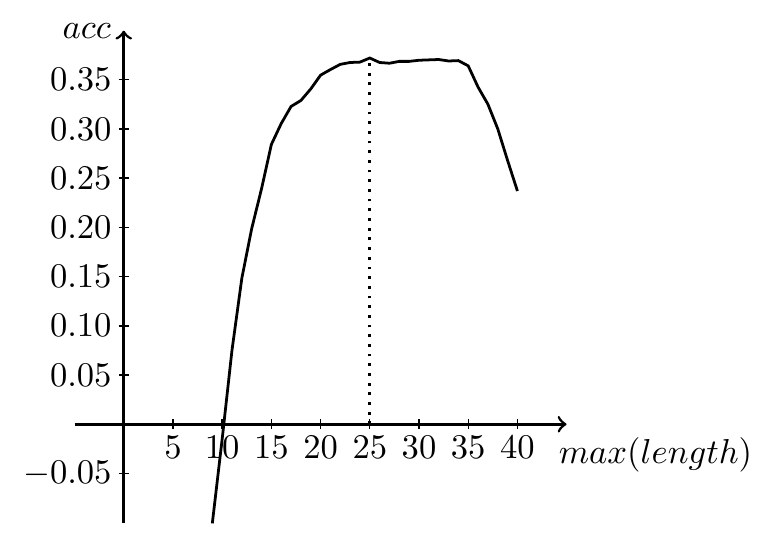}
	
	(a)
\end{minipage}

\begin{minipage}[b]{\columnwidth}
	\centering
	\includegraphics{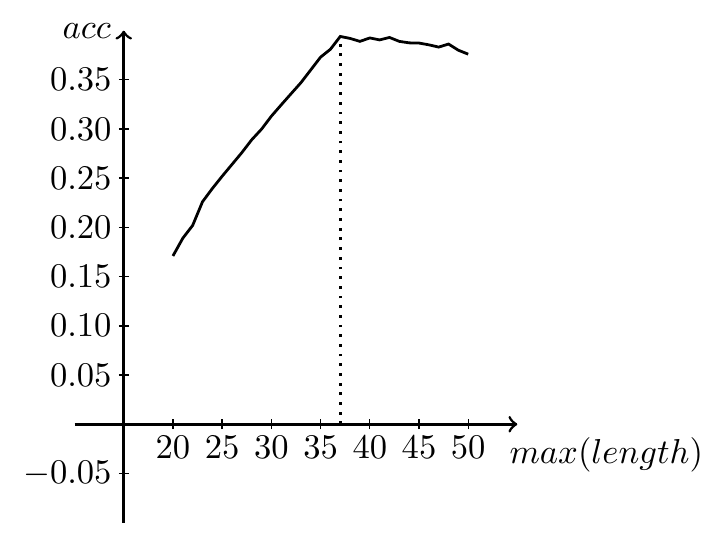}
	
	(b)
\end{minipage}

\caption[Cross-validation results for the maximum allowed length of sub-sequences.]{Cross-validation results for the maximum allowed length of sub-sequences. Figure (a) shows the cross-validation for the maximum length of single phonemes, figure (b) for the maximum length of biphones. The accuracy $acc$ measures the alignment to the correct phoneme sequence with formula \ref{eq_wer}.}
\label{fig_crossvallength}
\end{figure}

The minimum sequence length is set to 40~ms based on the shortest phonemes in the training data set. The maximum length is varied and the accuracy for each tested maximum length can be found in figure \ref{fig_crossvallength}. The highest accuracy is achieved with a maximum length of 25 frames (250~ms) for phonemes, and 37 frames (370~ms) for biphones. Accuracy values of phoneme sequences given the correct sequence can be calculated by formula \ref{eq_wer} in which $T$ stands for the total number of phonemes in the correct sequence, $C$ the number of correctly recognised phonemes, $S$ the number of substitutions, $D$ the number of deletions, and $I$ the number of insertions.

~\begin{equation} \label{eq_wer}
acc = \frac{T-S-I-D}{T} = \frac{C-I}{T}
\end{equation}~

Including information about all types of errors (subtitutions, deletion, insertions) gives the most meaningful performance measure. However, this requires to align the recognised and the correct sequence to find the respective counts. Aligning two sequences of different length in such a way that they maximise the accuracy measure of formula \ref{eq_wer} can be achieved with the Needleman-Wunsch algorithm \cite{Needleman1970}. This algorithm iteratively assigns error values to each alignment position resulting in an error value matrix for each pair of positions in both sequences. Afterwards, it backtracks the actual alignment by following the smallest error value from the end of the sequences back to the start.

Using the determined parameters, we can evaluate the performance on the final test set by trying to recognise the respective sequences using phonemes and biphones, solely phonemes, visemes and bi-visemes, or solely visemes. In line with the evaluations of others (e.g. \cite{Cappelletta2012}), we remove all short silent phonemes from the correct sequence before evaluating the accuracy. Only the silence at the beginning and the end of the sequence is allowed. When recognising phoneme sequences, the accuracy on the test set is $0.190$ when not using biphones and $0.201$ when using biphones. For the visemes, the test set accuracy is $0.372$ when not using bi-visemes and $0.394$ when using bi-visemes.

All in all, the continuous sequence recognition experiments allowed to find optimal parameter values for the maximum sub-sequence lengths. Using this parameters and all other parameters found by previous experiments, the test set could be classified with an accuracy of over $20\%$ for phonemes and almost $40\%$ for visemes. These results will be contrasted and discussed in the next section.

\section{Evaluation and Discussion}

In order to evaluate the achieved result, further analysis is required. The outcome itself needs to be scrutinised, looking at the strengths and weaknesses of the proposed method. Furthermore, the result needs to be compared to other existing methods based on the accuracy values reported in the literature. In the following, the first subsection will investigate the running time of the algorithm both theoretically and empirically. The second subsection will discuss the classification abilities concerning phonemes. Finally, the third subsection evaluates the viseme recognition performance.

\subsection{Run-Time Analysis}

The computational running time of an algorithm is an important factor to evaluate under which circumstances the method can be employed to solve further tasks. Especially for working with large datasets, an acceptable running time is vital. Additionally, it is relevant to determine if time-based tasks like visual speech recognition can be executed in real time which would open doors to new application possibilities. Only the running for the recognition of new unseen videos will be examined. The running time of the training process needs to be executed only once since the system uses general phonemes and does not need to be retrained.

The theoretical analysis of the running time shows that the pre-processing and mouth area segmentation has only a constant number of operations per video frame, leading to a running time in the order of  $\mathcal{O}(n)$, in which $n$ is the number of frames in the video. The Viterbi algorithm that runs on the entire series once has also a running time of $\mathcal{O}(n)$. The running time of creating the sub-sequences and extracting the features of each is determined by the running time of the feature extraction for one sub-sequence times the number of created sub-sequences. The number of sequences is $n\cdot d$ with $d$ being the number of allowed durations. Since $d$ is a constant factor, the number of sub-sequences is in the magnitude of $\mathcal{O}(n)$. The execution of the feature extraction and classification with the pre-trained SVM for one sub-sequence is again constant since all sub-sequences are of the same size. Therefore the running time of the entire feature extraction and classification is $\mathcal{O}(n)\cdot\mathcal{O}(1)=\mathcal{O}(n)$. The continuous sequence extraction as a final step always uses the same HMM to run the Viterbi algorithm using the extracted probabilities once. Hence, this step is determined by the Viterbi running time of $\mathcal{O}(n)$. All in all, all constituting steps have a linear running time leading to a total running time of $\mathcal{O}(n)$ as well.

\begin{figure}
\centering
\includegraphics{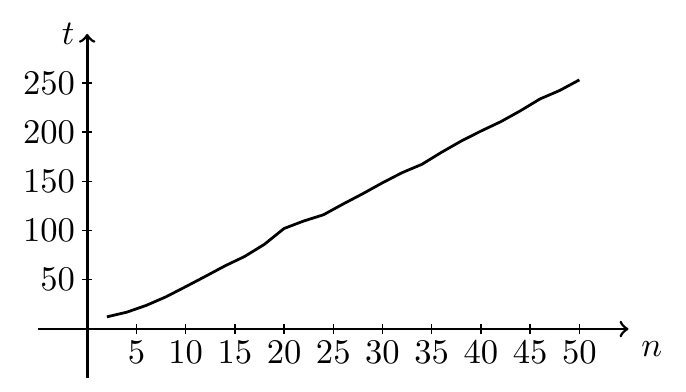}
\caption[Results of the empirical running time tests.]{Results of the empirical running time tests. The running time $t$ is shown in seconds depending on the number of frames $n$ in the input sequence.}
\label{fig_runtimetotal}
\end{figure}

Knowing that the algorithm has a linear execution time is a valuable information, as it implies that the running time for even very large datasets will not grow polynomially or exponentially, allowing the algorithm to run on any input in reasonable time. Moreover, in case the required time per frame is less than 40ms, the algorithm may run in real-time on continuous video inputs.  To check the actual behaviour, empirical tests of the running time for input sequences of different lengths can be conducted. Figure \ref{fig_runtimetotal} shows this evaluation for the execution time on a computer with an Intel\textsuperscript{\tiny\textregistered} Core\tiny\texttrademark\normalsize i5 CPU M 480 at 2.67GHz running on one core only and 3.7GB of memory. The resulting graph empirically confirms the linear running time. However, the slope of the curve implies that the average time per frame is 6.159 seconds which is far above the real-time condition of 40ms. This might be improved by an optimised code, faster computers, or the high parallelization capabilities of the algorithm, since almost every part of the process can be computed individually for each frame except the HMMs.

To further scrutinise the run-time weaknesses, we can look at the average time needed to execute each part of the algorithm. Table \ref{tab_runtime} shows a listing of the algorithm steps and the average time needed to complete the respective step. These values clearly show that the dominating part of the algorithm is the 3D-DCT feature extraction of each sub-sequence. As already mentioned, this might be easily improved by parallelizing the algorithm.  However, this remains a general disadvantage of 3D-feature-based approaches since all feasible selections of sub-sequences need to be examined, whereas 2D-feature-based approaches only need to calculate one set of features for each frame. 
\begin{table}
\resizebox{\columnwidth}{!}{
\centering
\begin{tabular}{lrr}
\toprule
Algorithm step & \multicolumn{1}{c}{$t$ in ms} & \multicolumn{1}{c}{\%}\\
\midrule
\textbf{1. Preprocessing} & \textbf{632.4} & \textbf{10.3\%} \\
\quad 1.1. Video loading & 11.1 & 0.2\% \\
\quad 1.2. Symmetry line detection & 294.1 & 4.8\% \\
\quad 1.3. Inner lower lip detection & 55.1 & 0.9\% \\
\quad 1.4. Mouth corner detection & 14.3 & 0.2\% \\
\quad 1.5. ROI extraction & 257.8 & 4.2\% \\
\textbf{2. Feature extraction} & \textbf{4577.2} & \textbf{74.2\%}\\
\quad 2.1. 3D-DCT calculation & 4577.2 & 74.2\%\\
\textbf{3. Classification} & \textbf{949.9} & \textbf{15.4\%}\\
\quad 3.1. SVM probability assignment & 511.9 & 8.3\%\\
\quad 3.2. HMM creation & 25.7 & 0.4\%\\
\quad 3.3. Continuous sequence extraction & 412.2 & 6.7\%\\
\bottomrule 
\end{tabular}
}
\caption[Listing of algorithm steps with their respective average running time per frame.]{Listing of algorithm steps with their respective average running time per frame. For each step, the running time $t$ in ms and the percentage of the full process is given. The three major steps written in bold display cumulative values.}
\label{tab_runtime}
\end{table}

\subsection{Phoneme Recognition}

This section will discuss the recognition abilities of the proposed method for phonemes. Firstly, the result itself will be analysed in more detail to see at which points errors occur, and to detect the strengths and weaknesses of the approach based on these findings. Secondly, the accuracy rates can be compared to other approaches in the literature.

\begin{table*}
\renewcommand{\arraystretch}{1.4}
\centering
\begin{sideways}
\resizebox{1.27\textwidth}{!}{

\begin{tabular}{|r||c|c|c|c|c|c|c|c|c|c|c|c|c|c|c|c|c|c|c|c|c|c|c|c|c|c|c|c|c|c|c|c|c|c|c|c|c|c|c|c||c|}
\hline
&\begin{turn}{-90}\textbf{V}\end{turn}&\begin{turn}{-90}\textbf{F}\end{turn}&
\begin{turn}{-90}\textbf{R}\end{turn}&\begin{turn}{-90}\textbf{UW}\end{turn}&
\begin{turn}{-90}\textbf{W}\end{turn}&\begin{turn}{-90}\textbf{ER}\end{turn}&
\begin{turn}{-90}\textbf{OW}\end{turn}&\begin{turn}{-90}\textbf{UH}\end{turn}&
\begin{turn}{-90}\textbf{M}\end{turn}&\begin{turn}{-90}\textbf{B}\end{turn}&
\begin{turn}{-90}\textbf{P}\end{turn}&\begin{turn}{-90}\textbf{AW}\end{turn}&
\begin{turn}{-90}\textbf{DH}\end{turn}&\begin{turn}{-90}\textbf{TH}\end{turn}&
\begin{turn}{-90}\textbf{SH}\end{turn}&\begin{turn}{-90}\textbf{CH}\end{turn}&
\begin{turn}{-90}\textbf{ZH}\end{turn}&\begin{turn}{-90}\textbf{JH}\end{turn}&
\begin{turn}{-90}\textbf{AO}\end{turn}&\begin{turn}{-90}\textbf{OY}\end{turn}&
\begin{turn}{-90}\textbf{S}\end{turn}&\begin{turn}{-90}\textbf{Z}\end{turn}&
\begin{turn}{-90}\textbf{IY}\end{turn}&\begin{turn}{-90}\textbf{AE}\end{turn}&
\begin{turn}{-90}\textbf{Y}\end{turn}&\begin{turn}{-90}\textbf{AA}\end{turn}&
\begin{turn}{-90}\textbf{IH}\end{turn}&\begin{turn}{-90}\textbf{AY}\end{turn}&
\begin{turn}{-90}\textbf{AH}\end{turn}&\begin{turn}{-90}\textbf{EH}\end{turn}&
\begin{turn}{-90}\textbf{EY}\end{turn}&\begin{turn}{-90}\textbf{D}\end{turn}&
\begin{turn}{-90}\textbf{T}\end{turn}&\begin{turn}{-90}\textbf{N}\end{turn}&
\begin{turn}{-90}\textbf{L}\end{turn}&\begin{turn}{-90}\textbf{K}\end{turn}&
\begin{turn}{-90}\textbf{G}\end{turn}&\begin{turn}{-90}\textbf{NG}\end{turn}&
\begin{turn}{-90}\textbf{HH}\end{turn}&\begin{turn}{-90}\textbf{sil}\end{turn}&\\
\hline\hline
\textbf{V}&\cellcolor{black!20}14&\cellcolor{black!20}&&&&&1&&&&&&1&&&&&&1&1&3&&3&2&1&&&&&&&1&&&2&&&&&&10\\\hline
\textbf{F}&\cellcolor{black!20}1&\cellcolor{black!20}4&&1&&&&&&&&&2&&&&&&1&1&&&&2&&1&1&&&&&&&1&&1&&&&&24\\\hline
\textbf{R}&6&2&\cellcolor{black!20}9&\cellcolor{black!20}3&\cellcolor{black!20}&\cellcolor{black!20}&\cellcolor{black!20}&\cellcolor{black!20}&&1&&&2&&3&&&&1&&6&1&8&8&&&1&&6&&&&&&9&5&&&3&&61\\\hline
\textbf{UW}&3&&\cellcolor{black!20}&\cellcolor{black!20}18&\cellcolor{black!20}&\cellcolor{black!20}&\cellcolor{black!20}&\cellcolor{black!20}&&&&&&&2&&&&7&&2&&&&&&&&4&&&&&&1&&&&&&18\\\hline
\textbf{W}&1&4&\cellcolor{black!20}3&\cellcolor{black!20}&\cellcolor{black!20}8&\cellcolor{black!20}&\cellcolor{black!20}1&\cellcolor{black!20}&&&&&1&&1&&&&3&&8&&3&7&1&&1&&2&&&1&&&&&&&&&21\\\hline
\textbf{ER}&&&\cellcolor{black!20}2&\cellcolor{black!20}5&\cellcolor{black!20}1&\cellcolor{black!20}&\cellcolor{black!20}&\cellcolor{black!20}&&&&&1&&1&&&&4&&9&&1&2&&&&&1&&&&1&1&3&&&&&&27\\\hline
\textbf{OW}&2&1&\cellcolor{black!20}&\cellcolor{black!20}1&\cellcolor{black!20}&\cellcolor{black!20}&\cellcolor{black!20}2&\cellcolor{black!20}&&&&&&&2&&&&2&&5&&3&4&&&1&&3&&&&1&&1&1&&&1&&11\\\hline
\textbf{UH}&&&\cellcolor{black!20}&\cellcolor{black!20}&\cellcolor{black!20}&\cellcolor{black!20}&\cellcolor{black!20}&\cellcolor{black!20}&&&&&&&&&&&&&1&&1&&&&&&&&&&&&&&&&&&7\\\hline
\textbf{M}&5&&1&2&&&4&&\cellcolor{black!20}6&\cellcolor{black!20}&\cellcolor{black!20}&&&&1&&&&2&&5&&3&3&1&&&&1&&&2&&&3&&&&&&35\\\hline
\textbf{B}&&&&&&&&&\cellcolor{black!20}&\cellcolor{black!20}&\cellcolor{black!20}&&&&1&&&&1&&5&&5&1&&&2&&1&&&2&&&1&&&&&&22\\\hline
\textbf{P}&5&&&1&1&&&&\cellcolor{black!20}&\cellcolor{black!20}&\cellcolor{black!20}&&&&&&&&1&&7&&3&5&&&1&&2&2&&&&&3&&&&&&26\\\hline
\textbf{AW}& & && &&&&&& &&\cellcolor{black!20}&&&&&&&&&&&&&&&&&&&&&&&&&&&&&7\\\hline
\textbf{DH}&1&1&&2&&&&&&1&&&\cellcolor{black!20}3&\cellcolor{black!20}&1&&&&4&&4&&5&8&2&&&&&&&&&&1&1&&&&&29\\\hline
\textbf{TH}& & && &&&&&&&&&\cellcolor{black!20}&\cellcolor{black!20}& &&&&&&1&&&2&&&&&&&&&&&1&&&&&&5\\\hline
\textbf{SH}&1&1&&1&&&&&&&&&&&\cellcolor{black!20}9&\cellcolor{black!20}&\cellcolor{black!20}&\cellcolor{black!20}&1&&3&&&1&1&&&&1&&&&&&1&1&&&&&17\\\hline
\textbf{CH}&1&&& &&&&&&&&& &&\cellcolor{black!20}&\cellcolor{black!20}&\cellcolor{black!20}&\cellcolor{black!20}& & &1&&&&&&&&&&&&&&2&&&&&&6\\\hline
\textbf{ZH}& &&& &&&&&&&&& &&\cellcolor{black!20}&\cellcolor{black!20}&\cellcolor{black!20}&\cellcolor{black!20}&1& & &&&&&&&&&&&&&&&&&&&&4\\\hline
\textbf{JH}& &&&1&&&&&&&&&1&&\cellcolor{black!20}&\cellcolor{black!20}&\cellcolor{black!20}&\cellcolor{black!20}& & &1&&1&2&1&&&&1&&&&&&1&&&&&&7\\\hline
\textbf{AO}&1&&&4&&&&&&&&&2&&1&&&&\cellcolor{black!20}21&\cellcolor{black!20}&6&&4&2&&&&&2&&&1&&&&&&&&&12\\\hline
\textbf{OY}&2&&&&&&&&&&&&&&&&&&\cellcolor{black!20}3&\cellcolor{black!20}&3&&&1&&&&&&&&&&&&&&&&&3\\\hline
\textbf{S}&6&&&2&&&1&&&&&&&&2&&&&1&&\cellcolor{black!20}86&\cellcolor{black!20}&1&1&&&1&&4&&&1&1&&1&&&&&&22\\\hline
\textbf{Z}&1&&&2&&&1&&&&&&&&1&&&2&1&&\cellcolor{black!20}9&\cellcolor{black!20}2&6&2&&&&&3&1&&3&&&1&2&&&&&25\\\hline
\textbf{IY}&6&&&1&&&3&&&&&&4&&1&&&&3&&6&&\cellcolor{black!20}56&\cellcolor{black!20}6&\cellcolor{black!20}&\cellcolor{black!20}&\cellcolor{black!20}2&\cellcolor{black!20}&\cellcolor{black!20}4&\cellcolor{black!20}&\cellcolor{black!20}&2&&&1&2&&&&&39\\\hline
\textbf{AE}&1&&&&&&&&1&&&&1&&2&&&&4&&7&&\cellcolor{black!20}4&\cellcolor{black!20}43&\cellcolor{black!20}&\cellcolor{black!20}&\cellcolor{black!20}1&\cellcolor{black!20}&\cellcolor{black!20}4&\cellcolor{black!20}&\cellcolor{black!20}&&&&1&&&&1&&22\\\hline
\textbf{Y}&&&&&&&&&1&&&&1&&3&&&&1&&1&1&\cellcolor{black!20}3&\cellcolor{black!20}3&\cellcolor{black!20}4&\cellcolor{black!20}&\cellcolor{black!20}1&\cellcolor{black!20}&\cellcolor{black!20}1&\cellcolor{black!20}&\cellcolor{black!20}&&&&&&&&2&&17\\\hline
\textbf{AA}&3&&1&7&&&&&1&&&&&1&2&&&&2&&4&&\cellcolor{black!20}2&\cellcolor{black!20}2&\cellcolor{black!20}&\cellcolor{black!20}&\cellcolor{black!20}&\cellcolor{black!20}&\cellcolor{black!20}2&\cellcolor{black!20}&\cellcolor{black!20}&&&&1&&&&&&33\\\hline
\textbf{IH}&1&1&1&2&&&&&1&&&&1&&&&&&5&&11&&\cellcolor{black!20}10&\cellcolor{black!20}6&\cellcolor{black!20}&\cellcolor{black!20}&\cellcolor{black!20}19&\cellcolor{black!20}&\cellcolor{black!20}2&\cellcolor{black!20}1&\cellcolor{black!20}&&&&2&2&&&1&&63\\\hline
\textbf{AY}&1&&&3&&&2&&&&&&&&1&&&&1&&3&&\cellcolor{black!20}2&\cellcolor{black!20}6&\cellcolor{black!20}2&\cellcolor{black!20}&\cellcolor{black!20}1&\cellcolor{black!20}&\cellcolor{black!20}&\cellcolor{black!20}&\cellcolor{black!20}&&1&&&2&&&&&14\\\hline
\textbf{AH}&&1&&4&1&&1&&&&&&1&&6&&&&8&&14&&\cellcolor{black!20}10&\cellcolor{black!20}8&\cellcolor{black!20}2&\cellcolor{black!20}&\cellcolor{black!20}2&\cellcolor{black!20}1&\cellcolor{black!20}58&\cellcolor{black!20}1&\cellcolor{black!20}&3&&&2&5&&&3&&92\\\hline
\textbf{EH}&&&1&3&&&&&&&&&&&1&&&&2&&4&1&\cellcolor{black!20}3&\cellcolor{black!20}4&\cellcolor{black!20}&\cellcolor{black!20}&\cellcolor{black!20}&\cellcolor{black!20}&\cellcolor{black!20}1&\cellcolor{black!20}1&\cellcolor{black!20}&&1&&2&&&&1&&25\\\hline
\textbf{EY}&1&&&2&&&1&&&&&&2&&&&&&4&&6&&\cellcolor{black!20}&\cellcolor{black!20}4&\cellcolor{black!20}&\cellcolor{black!20}&\cellcolor{black!20}&\cellcolor{black!20}&\cellcolor{black!20}&\cellcolor{black!20}&\cellcolor{black!20}&1&&&2&&&&&&15\\\hline
\textbf{D}&&&&&1&&&&2&&&&3&&7&&&&1&&15&1&8&5&1&&1&&3&1&&\cellcolor{black!20}9&\cellcolor{black!20}&\cellcolor{black!20}&\cellcolor{black!20}3&&&&&&35\\\hline
\textbf{T}&7&1&2&7&1&&1&&1&1&1&&4&&3&&&&2&&16&1&4&9&5&&3&&2&&&\cellcolor{black!20}2&\cellcolor{black!20}4&\cellcolor{black!20}&\cellcolor{black!20}4&3&&&1&&85\\\hline
\textbf{N}&9&&1&1&&&&&&2&&&3&&2&&&1&1&&12&&6&6&&&&&5&&&\cellcolor{black!20}2&\cellcolor{black!20}1&\cellcolor{black!20}6&\cellcolor{black!20}8&3&1&&&&71\\\hline
\textbf{L}&12&&&3&1&&1&&1&1&&&&&3&&&&&&7&&5&6&&&2&&2&&&\cellcolor{black!20}&\cellcolor{black!20}1&\cellcolor{black!20}1&\cellcolor{black!20}35&&&&1&&45\\\hline
\textbf{K}&5&&&1&1&1&1&&&&&&1&&5&&&&2&&6&1&2&6&&&1&&4&1&&&&&5&\cellcolor{black!20}8&\cellcolor{black!20}&\cellcolor{black!20}&&&52\\\hline
\textbf{G}&1&&&&&&&&&&&&3&&2&&&&2&&4&&3&1&&&1&&1&&&&1&&1&\cellcolor{black!20}&\cellcolor{black!20}&\cellcolor{black!20}&&&20\\\hline
\textbf{NG}&1&&&&&&&&&&&&&&1&&&&&&3&&1&&&&&&&&&1&1&&2&\cellcolor{black!20}1&\cellcolor{black!20}&\cellcolor{black!20}&&&10\\\hline
\textbf{HH}&1&&1&&&&2&&&&&&1&&4&&&&&&2&&&1&1&&&&6&&&&&&2&&&&\cellcolor{black!20}2&&15\\\hline
\textbf{sil}&&&&&&&&&&&&&&&&&&&&&&&&&&&&&&&&&&&&&&&&\cellcolor{black!20}160&\\\hline
\hline
\textbf{}&8&&&1&1&&1&&&&&&2&&1&&&&&&6&1&4&5&&&1&&1&&1&2&&1&1&2&&&&&\\\hline
\end{tabular}
}
\end{sideways}

\caption{Confusion matrix for phonemes. Rows represent the correct phoneme, whereas columns represent the actual output of the classifier. The separate column at the right shows the deletion errors, the separate row at the bottom shows insertion errors. Shaded areas highlight phoneme pairs of the same viseme. For clarity, zeros are left blank.}

\label{tab_confphon}

\end{table*}

In order to scrutinize the source of the classification errors, it is helpful to consult the confusion matrix of the result in table \ref{tab_confphon}. This matrix disassembles the accuracy into values for each single phoneme. Each row in the matrix corresponds to the correct phoneme which was to be recognised. The columns represent the actual output of the classifier. Classifications on the diagonal of the matrix are correct, all other counts represent substitution errors. The deletion errors for each phoneme are shown in the last separate column and the insertion errors in the last separate row. The phonemes are ordered in such a way that phonemes which can be grouped to visemes are in adjacent rows. This is to highlight the difference between intra-viseme substitutions, which are likely since the visemes should be visually indistinguishable from one another, and inter-viseme substitutions.

In total, of the 2\,728 phonemes that were to be recognised in the test set, 587 were correct, 1\,089 were substituted (of which 156 are intra-viseme substitutions), 1\,052 were deleted, and 39 phonemes were inserted additionally. This result clearly shows two main sources of errors: Firstly, the proposed method tends to produce shorter sequences than the correct ones, leading to a high number of deletions and only a few insertions. Secondly, it can be observed that the recognition is biased towards frequent phonemes leading to substitutions in which frequent phonemes like S, AE or AO are recognised although they are not present and infrequent phonemes like AW, CH, ZH are not even recognised when they are there. The first type of error is most likely based on the problem that single long sub-sequences with a high probability are preferred to multiple shorter sequences with mixed likelihoods. A possible countermeasure could be to penalise long sequences by weighting the probabilities $p$ of sub-sequences of length $l$ with a function similar to $p' = p\cdot\exp(-\alpha l)$ in which the parameter $\alpha$ is still to be determined. The second type of error arises due to the unbalanced numbers of training samples of each phoneme. Phonemes with many training samples get higher probabilities as this leads to higher accuracy values on the cross-validation set. In order to counteract this problem more training samples of the rare phonemes need to be collected to have equal numbers of each. Alternatively, the cross-validation can try to optimise the average relative accuracy of each individual phoneme and not the total proportion of correctly classified phonemes. However, counteracting this problem will reduce the accuracy on real test sets as they will contain more of the frequent phonemes and classifiers with bias towards frequent phonemes will perform better. A final observation from the confusion matrix is that intra-viseme substitution errors are not as high as expected. This means that visual speech recognition can classify phonemes of the same viseme better than chance (especially for visemes /A, /B, /I, and /J) which implies that the linguistic viseme models cluster phonemes although they are not visually equal in every aspect. 

\begin{figure}
\centering

\begin{minipage}[b]{\columnwidth}
\centering
	\includegraphics{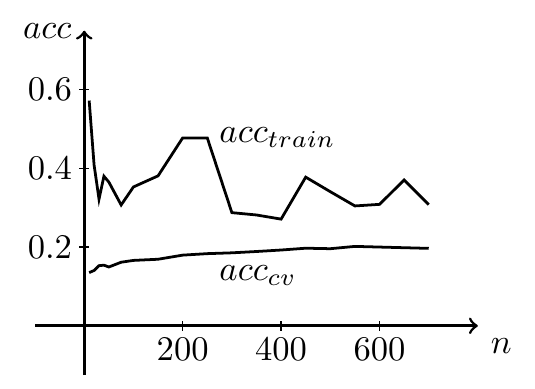}

(a)
\end{minipage}

\begin{minipage}[b]{\columnwidth}
\centering
	\includegraphics{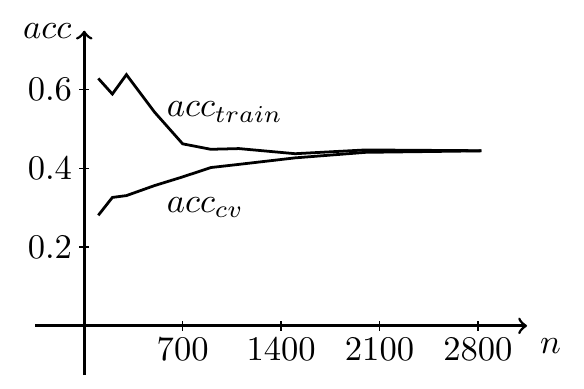}

(b)
\end{minipage}

\caption[Learning curves for phoneme and viseme classifiers.]{Learning curves for phoneme (a) and viseme (b) classifiers. Each figure shows the accuracy on the training set $acc_{train}$ and on the cross-validation set $acc_{cv}$ related to the number of training samples $n$ of each phoneme or viseme used for training.}
\label{fig_learningcurves}
\end{figure}

In general, the total error of a trained model can be ascribed to three components: Firstly, to the ambiguity of the task itself which is unexplainable by any model. Secondly, to flaws in the model which cannot predict the correct output even under perfect training conditions as the model is not able to fully describe the problem at hand. Thirdly, to errors based on imperfect training conditions like too few training samples which are not enough to find the best model. Looking at the `learning curves' of the training process can give hints on whether the result suffers from the third type of error and consequently whether more training samples are likely to improve the result or not. The `learning curves' relate the accuracy values to the number of training samples used to train the model. The accuracy of the training set is usually decreasing the more samples are used since the model cannot fit many data points as perfect as it can fit few data points only. On the other hand, the accuracy on the cross-validation set is usually increasing the more samples are used for training as the model becomes more generic and therefore better at classifying the unseen samples from the cross-validation set. Both accuracies, of the training and the cross-validation set, tend to converge to one value as soon as enough training samples are available. Thus, the learning curves can be used to determine if more training samples are likely to improve the result. Figure \ref{fig_learningcurves}a shows the learning curves for the phoneme training process, in which the accuracy denotes the proportion of correctly classified phonemes in the training or cross-validation set respectively. The curves still have a gap of approximately 10\% in accuracy, rendering it very likely that more training samples will improve the result considerably. Still, there will be a large remaining error that is either based on the ambiguity of the problem or the insufficient ability of the model to describe the mouth movements. Since trained human lip-readers outperform current automatic systems \cite{Lan2010}, it can be assumed that a high model insufficiency remains.

Finally, the achieved results can be compared to other reported results in the literature. Care has to be taken, since many authors restricted their studies on specific phoneme sets or specific phoneme combinations, e.g. Werda et al. \cite{Werda2007}, rendering their results unsuitable for comparison. Table \ref{tab_resultsphon} shows accuracy values according to formula \ref{eq_wer} of our approach and other reported approaches which tried to recognise arbitrary phoneme sequences in the English language of any phoneme combination. Newman et al. \cite{Newman2010} use the MOCHA-TIMIT database which includes additional measurements of the mouth to allow better recognition results. Using the frontal camera image only and extracting shape-based features from an AAM lip segmentation, they achieve an accuracy of 14.5\%, claiming that this is the upper bound that will be reached for video only recognition. On the other hand, Ramage \cite{Ramage2011} uses a 2D-DCT feature extractor together with the respective first and second derivatives to test the accuracy on the VidTimit database as well. He achieves 18.6\% accuracy using biphones and even triphones to train HMMs for the continuous sequence extraction. In line with our finding that visemes do not necessarily cluster phonemes sensibly, he argues that any viseme clustering will reduce the recognition ability on future tasks.

It can be observed that the proposed method is currently the best among comparable approaches, with the basic version of not using biphones being already better than all other other approaches. Using biphones improves the result even further. A one-tailed t-test contrasting our results with those of Ramage \cite{Ramage2011} shows that the improvement from 18.6\% to 20.1\% using 80 test sentences is statistically significant with a probability of 0.0112 of getting the same result by chance alone ($t=2.302$). We can therefore conclude that the usage of 3D features improves the result compared to 2D features.

Moreover, a one-tailed paired t-test between the accuracy of our method with and without biphones reveals that the improvement due to biphones from 19.0\% to 20.1\% is marginally significant as well with a probability of 0.0571 of achieving the same improvement by chance ($t=1.597$).
This observation shows that taking care of the co-articulation effects by using biphones or triphones improves the result considerably. This also implies that the introduction of triphones in our model might result in further improvements. However, this requires additional training samples since the VidTimit database does not even provide enough examples to train the single phoneme classifier without a large remaining gap in the cross-validation performance as shown by the learning curves.

\begin{table}
\resizebox{\columnwidth}{!}{
\centering
\begin{tabular}{llclr}
\toprule
reference & features & B/T & database & acc. \\
\midrule
Newman et al. \cite{Newman2010} & AAM & & MOCHA & 14.5\%  \\
Ramage \cite{Ramage2011} & 2D-DCT & X & VidTimit & 18.6\% \\
our approach & 3D-DCT &  & VidTimit & \textbf{19.0\%}  \\
our approach & 3D-DCT & X & VidTimit & \textbf{20.1\%}  \\
\bottomrule
\end{tabular}
}
\caption[Comparison of accuracy values of the proposed method and other phoneme recognition methods.]{Comparison of accuracy values of the proposed method and other phoneme recognition methods. The columns indicate the reference, the feature extraction approach, the usage of biphones or triphones (B/T), the used database, and the achieved accuracy (acc.). Accuracies of our approach are in bold.}
\label{tab_resultsphon}
\end{table}

\subsection{Viseme Recognition}

Besides phoneme recognition performance, the proposed algorithm can also be analysed in terms of viseme recognition performance. The same procedure can be applied to find strengths and weaknesses revealed by the result and to compare the outcome with findings of other authors.

\begin{table*}
\centering
\begin{tabular}{|l||c|c|c|c|c|c|c|c|c|c|c|c||c|}
\hline
&\textbf{/A}&\textbf{/B}&\textbf{/C}&\textbf{/D}&\textbf{/E}&\textbf{/F}&\textbf{/G}&
\textbf{/H}&\textbf{/I}&\textbf{/J}&\textbf{/K}&\textbf{/L}&\\
\hline\hline
\textbf{/A}&5&1&3&&6&1&&7&2&8&1&&46\\\hline
\textbf{/B}&2&112&1&&11&6&3&12&13&7&5&&157\\\hline
\textbf{/C}&2&5&56&&11&2&1&11&3&1&&&73\\\hline
\textbf{/D}&&&&&&&1&1&1&&&&4\\\hline
\textbf{/E}&&1&&&30&&&5&&&2&&34\\\hline
\textbf{/F}&&2&&&6&6&2&7&&7&1&&38\\\hline
\textbf{/G}&&2&&&2&&20&4&1&1&&&38\\\hline
\textbf{/H}&1&1&1&&5&4&6&104&3&4&2&&58\\\hline
\textbf{/I}&&8&4&&15&3&23&21&387&2&6&&273\\\hline
\textbf{/J}&5&9&8&&35&6&7&34&1&122&12&&238\\\hline
\textbf{/K}&1&4&1&&6&2&8&11&3&5&27&&92\\\hline
\textbf{/L}&&&&&&&&&&&&160&\\
\hline\hline
&&3&1&&7&1&4&10&2&5&4&&\\\hline
\end{tabular}

\caption[Confusion matrix for viseme recognition.]{Confusion matrix for viseme recognition. Rows represent the correct viseme, whereas columns represent the actual output of the classifier. The separate column at the right shows the deletion errors, the separate row at the bottom shows insertion errors. For clarity, zeros are left blank.}
\label{tab_confvis}
\end{table*}

\begin{table*}
\centering
\begin{tabular}{lllr}
\toprule
reference & feature extraction & database & accuracy \\
\midrule
Cappelletta and Harte \cite{Cappelletta2012} & 2D-PCA, with 3-state HMMs & VidTimit & 29-30\%  \\
Cappelletta and Harte \cite{Cappelletta2012} & 2D-PCA, with 4-state HMMs & VidTimit & 34-35\%  \\
Cappelletta and Harte \cite{Cappelletta2012} & Optical flow, with 3-state HMMs & VidTimit & 26-27\%  \\
Cappelletta and Harte \cite{Cappelletta2012} & Optical flow, with 4-state HMMs & VidTimit & 33-34\%  \\
Cappelletta and Harte \cite{Cappelletta2012} & 2D-DCT, with 3-state HMMs & VidTimit & 37-38\%  \\
Cappelletta and Harte \cite{Cappelletta2012} & 2D-DCT, with 4-state HMMs & VidTimit & 33-34\%  \\
our approach & 3D-DCT (visemes only) & VidTimit & \textbf{37.2\%}  \\
our approach & 3D-DCT (visemes + bi-visemes) & VidTimit & \textbf{39.4\%}  \\
\bottomrule
\end{tabular}
\caption[Comparison of accuracy values of the proposed method and other viseme recognition methods.]{Comparison of accuracy values of the proposed method and other viseme recognition methods. The columns indicate the reference, the feature extraction approach, the used database, and the achieved accuracy. Accuracies of our approach are in bold.}
\label{tab_resultsvis}
\end{table*}

Table \ref{tab_confvis} shows the confusion matrix for the visemes of the Jeffers mapping. Of the 2\,518 visemes in the correct transcription, 1\,029 were identified correctly, 438 were substituted, 1\,051 were deleted and 37 were inserted additionally. Not surprisingly, the same two types of errors as for phonemes can be observed: Firstly, the high number of deletions due to the problem that recognised sequences tend to be shorter than the correct transcription. Secondly, the bias towards frequent visemes like /H which get more detections than actually present, whereas infrequent visemes like /D get no detection at all. Identical countermeasures can be taken as for the phonemes with similar expected changes. The probabilities can be weighted to avoid the preference for long visemes and the training can work with balanced training sets to reduce the impact of the second problem. Again, it might not be the intention to counteract the bias towards frequent visemes as it will reduce the accuracy on new real data sets.

The second experiment to be conducted is the creation of the learning curves to check whether more training data will improve the result. Figure \ref{fig_learningcurves}b shows the learning curves for the viseme training. In contrast to the findings of the phoneme training, these curves clearly show that the training accuracy and the cross-validation accuracy have converged, leading to the conclusion that more training data will almost certainly not help to improve the result. The remaining error is therefore only based on model flaws and the inherent problem ambiguity. Again, it can be safely assumed that better models exist since the performance of human lip-readers is higher than the performance of computers \cite{Lan2010}, meaning that the remaining error is not exclusively based on the problem ambiguity.

When comparing the results to those found in the literature, it is again necessary to exclude papers which did not use the same phoneme-viseme mapping or tested the accuracy under different conditions. Available accuracy values are summarised in table \ref{tab_resultsvis}. All other approaches are taken from Cappelletta and Harte  \cite{Cappelletta2012} who compared DCT, PCA and optical flow feature extraction methods by checking the accuracy on the Jeffers visemes in the VidTimit database. Unfortunately, the exact numbers are not given in the paper and can only be estimated from graphs showing the performance of different approaches. Their best approach, 2D-DCT features with 3-state HMMs for continuous sequence extraction, performs equally well as our method when trained with visemes alone. Adding bi-visemes to the training process improves the result by about 2\% to 39.4\% which is clearly better than all other methods. It is interesting to note, however, that the approach of Cappelletta and Harte suffers from the exact opposite problem of having very few deletions and a large number of insertions. In order to check whether the found improvements are statistically significant, t-tests based on the 80 test sentences were conducted. A one-tailed t-test contrasting our results with that of Cappelletta and Harte \cite{Cappelletta2012}, assuming an accuracy of their method of 37.5\%, shows that the improvement to 39.4\% is statistically significant with a probability of 0.0078 of getting the same result by chance alone ($t=2.472$). Moreover, a one-tailed paired t-test between the accuracy of our method with and without biphones reveals that the improvement due to biphones from 37.2\% to 39.4\% is highly significant as well with a probability of 0.0072 of achieving the same improvement by chance ($t=2.500$).

In summary, our approach again outperforms all other methods tested under comparable settings and conditions. The viseme classification suffers from the same problems as the phoneme classification. However, the number of training samples was sufficient to train the model and more data will not improve the result. Yet, further improvements could be achieved by overcoming the problem of preferring long visemes in the continuous sequence extraction. 

\section{Conclusion}

In order to judge the method and the result in the context of the current research, we can look at design novelties and design improvements utilised by the approach leading to three specific contributions of this work. Based upon open questions and missing links identified by the literature review, our approach has three special design features. Firstly, it combines the robustness and accuracy of model-based lip segmentation methods with the fast computation and simplicity of a region-based output. It also uses a robust tracking method by applying HMMs instead of classic forward-only tracking methods. Secondly, it uses 3D features for continuous speech to include time-dependant information which is lost with 2D features. This approach 
has not been tried previously, mostly due to the lack of information about the positions and durations of phonemes. Our method overcomes the problem by using a fast feature extraction that allows to extract features of all feasible sub-sequences to combine the classification results afterwards with a tailor-made HMM. Thirdly, the review showed that human speech has two aggravating properties -- co-articulation and audio-visual time shift -- which impede the visual speech recognition. Unlike many other authors, our approach actively counteracts both problems. These three special design features are well-suited in the current research as they continue and improve already explored methods and add new combinations of solutions that were shown to work well. The design is also based on the experience and problems other authors encountered in their investigations. As a consequence, the proposed method and its result makes three specific contributions to the research area:
\begin{enumerate}
\item The method shows that the problem of applying 3D features in continuous sequence recognition can be solved despite the unknown phoneme positions and durations by calculating probabilities for each phoneme at each starting point and for each duration. The actual sequence can be extracted afterwards by applying a tailor-made Hidden Markov Model.
\item The current paper revealed that 3D features perform better than 2D features on the same dataset and under the same test conditions. This finding holds true for phonemes as well as visemes.
\item The analysis of the result showed that the usage of biphones has a considerably positive effect on the accuracy. Although biphones and triphones have been used by others, none explicitly compared the accuracy using an otherwise identical method with and without using biphones.
\end{enumerate}

The last part of the conclusion deals with possible improvements and application areas. The evaluation and analysis of the results uncovered some problems of the approach which need further attention: The execution is slower than other methods, especially due to the 3D feature extraction which works on all eligible sub-sequences and therefore takes much time. The result has a high number of deletion errors based on the preference for long phonemes and there is a bias towards frequent phonemes. Moreover, the SVM models for the phoneme recognition had not enough training data to release their full potential. Consequently, the result might be improved by using more training data. The AV-TIMIT database is a likely candidate as it provides similar input data as the VidTimit database but is approximately 10 times larger. However, legal reasons prevent the usage outside the Massachusetts Institute of Technology. Another improvement, tackling the long phoneme preference could be a weighting function which gives shorter phonemes a higher probability. Furthermore, the high running time can be reduced by applying parallelization and other code optimisations. Having more training samples, the accuracy can likely be improved by adding triphones to the process. Additionally, language and grammar models can be applied to post-process the result and to ensure only valid English words and sentences are extracted or at least to ensure that only likely transitions from one phoneme to the next are used. Even further extensions could include other video poses like side views or complications like non-language expressions and silent sections.

\footnotesize
\bibliographystyle{plain}

\begin{thebibliography}{10}
\providecommand{\url}[1]{{#1}}
\providecommand{\urlprefix}{URL }
\expandafter\ifx\csname urlstyle\endcsname\relax
  \providecommand{\doi}[1]{DOI~\discretionary{}{}{}#1}\else
  \providecommand{\doi}{DOI~\discretionary{}{}{}\begingroup
  \urlstyle{rm}\Url}\fi

\bibitem{Abel2009}
Abel, A., Hussain, A., Nguyen, Q.D., Ringeval, F., Chetouani, M., Milgram, M.:
  Maximising audiovisual correlation with automatic lip tracking and vowel
  based segmentation.
\newblock In: Proc. of the jt. COST 2101 and 2102 Int. Conf. on Biom. ID
  Management and Multimodal Commun., pp. 65--72 (2009)

\bibitem{Aleksic2004}
Aleksic, P.S., Katsaggelos, A.K.: Comparison of low- and high-level visual
  features for audio-visual continuous automatic speech recognition.
\newblock In: Proc. of the IEEE Int. Conf. on Acoust., Speech, and Signal
  Process., pp. 917--920 (2004)

\bibitem{Aronoff2003}
Aronoff, M., Rees-Miller, J.: The handbook of linguistics, vol.~43.
\newblock Wiley-Blackwell (2003)

\bibitem{Cappelletta2010}
Cappelletta, L., Harte, N.: Nostril detection for robust mouth tracking.
\newblock In: Proc. of the IET Ir. Signals and Syst. Conf., pp. 239--244 (2010)

\bibitem{Cappelletta2012}
Cappelletta, L., Harte, N.: Phoneme-to-viseme mapping for visual speech
  recognition.
\newblock In: Proc. of the Int. Conf. on Pattern Recognit. Appl. and Methods
  (2012)

\bibitem{Castaneda2005}
Casta{\~n}eda, B., Cockburn, J.C.: Reduced support vector machines applied to
  real-time face tracking.
\newblock In: Proc. of the IEEE Int. Conf. on Acoust., Speech, and Signal
  Process., vol.~2, pp. 673--676 (2005)

\bibitem{Chang2011}
Chang, C.C., Lin, C.J.: {LIBSVM}: A library for support vector machines.
\newblock ACM Trans. on Intell. Syst. and Technol. \textbf{2}, 27:1--27:27
  (2011).
\newblock Software available at \url{http://www.csie.ntu.edu.tw/~cjlin/libsvm}

\bibitem{Chitu2010}
Chiţu, A.G., Driel, K., Rothkrantz, L.J.M.: Automatic lip reading in the dutch
  language using active appearance models on high speed recordings.
\newblock In: Proc. of the 13th Int. Conf. on Text, Speech and Dialogue,
  TSD'10, pp. 259--266. Springer-Verlag, Berlin, Heidelberg (2010)

\bibitem{Chitu2007}
Chiţu, A.G., Rothkrantz, L.J.M., Wiggers, P., Wojdel, J.C.: Comparison between
  different feature extraction techniques for audio-visual speech recognition.
\newblock J. on Multimodal User Interfaces \textbf{1}(1), 7--20 (2007)

\bibitem{Deypir2011}
Deypir, M., Alizadeh, S., Zoughi, T., Boostani, R.: Boosting a multi-linear
  classifier with application to visual lip reading.
\newblock Expert Syst. Appl. \textbf{38}(1), 941--948 (2011)

\bibitem{Duchnowski1995}
Duchnowski, P., Hunke, M., Busching, D., Meier, U., Waibel, A.: Toward
  movement-invariant automatic lip-reading and speech recognition.
\newblock In: Proc. of the IEEE Int. Conf. on Acoustics, Speech, and Signal
  Process., vol.~1, pp. 109--112 (1995)

\bibitem{Estellers2012}
Estellers, V., Thiran, J.P.: Multi-pose lipreading and audio-visual speech
  recognition.
\newblock EURASIP J. on Adv. in Signal Process. \textbf{2012}(1), 1--23 (2012)

\bibitem{Eveno2004}
Eveno, N., Caplier, A., Coulon, P.Y.: Accurate and quasi-automatic lip
  tracking.
\newblock IEEE Trans. on Circuits and Syst. for Video Technol. \textbf{14}(5),
  706--715 (2004)

\bibitem{Garcia2009}
Garcia, D.: dctn - {N}-dimensional {Discrete Cosine Transform}.
\newblock \url{http://www.biomecardio.com/ matlab/dctn.html} (2009).
\newblock Accessed 15 Jun 2013

\bibitem{Gorman2007}
Gorman, K.: Automatic speech segmentation with {HTK}.
\newblock \url{http://www.ling.upenn.edu/ kgorman/papers/
  segmentation/.speechseg.html} (2007).
\newblock Accessed 28th Feb 2013

\bibitem{Hassanat2011}
Hassanat, A.: Visual speech recognition.
\newblock In: I.~Ipsic (ed.) Speech and Lang. Technol. InTech (2011)

\bibitem{Hassanat2010}
Hassanat, A., Jassim, S.: Color-based lip localization method.
\newblock In: SPIE Def., Security, and Sens., p. 77080Y. Int. Soc. for Optics
  and Photonics (2010)

\bibitem{Hazen2004}
Hazen, T.J., Saenko, K., La, C.H., Glass, J.R.: A segment-based audio-visual
  speech recognizer: data collection, development, and initial experiments.
\newblock In: Proc. of the 6th Int. Conf. on Multimodal Interfaces, pp.
  235--242 (2004)

\bibitem{Hosseini2009}
Hosseini, M.M., Ghofrani, S.: Automatic lip extraction based on wavelet
  transform.
\newblock In: WRI Glob. Congres. on Intell. Syst., vol.~4, pp. 393--396 (2009)

\bibitem{Huang2010}
Huang, Y.H., Liang, J., Pan, A.C., Fan, X.Y.: A new lip-automatic detection and
  location algorithm in lip-reading system.
\newblock In: Proc. of the IEEE Int. Conf. on Syst. Man and Cybern., pp.
  2402--2405 (2010)

\bibitem{Ibrahim2012}
Ibrahim, M., Mulvaney, D.J.: Geometry based lip reading system using multi
  dimension dynamic time warping.
\newblock In: Vis. Comm. and Image Process., pp. 1--6 (2012)

\bibitem{Jeffers1971}
Jeffers, J., Barley, M.: Speechreading (lipreading).
\newblock Thomas Springfield (1971)

\bibitem{Jiang2006}
Jiang, M., Gan, Z., He, G., Gao, W.: Combining particle filter and active shape
  models for lip tracking.
\newblock In: Sixth World Congres. on Intell. Control and Autom., vol.~2, pp.
  9897--9901 (2006)

\bibitem{Johnson2005}
Johnson, M.: Capacity and complexity of {HMM} duration modeling techniques.
\newblock IEEE Trans. on Signal Process. Lett. \textbf{12}(5), 407--410 (2005)

\bibitem{Karlsson2012}
Karlsson, S.M., Bigun, J.: Lip-motion events analysis and lip segmentation
  using optical flow.
\newblock In: IEEE Comput. Soc. Conf. on Comput. Vis. and Pattern Recognit.
  Workshops, pp. 138--145 (2012)

\bibitem{Kaucic1996}
Kaucic, R., Dalton, B., Blake, A.: Real-time lip tracking for audio-visual
  speech recognition applications.
\newblock In: B.~Buxton, R.~Cipolla (eds.) Computer Vision — ECCV '96,
  \emph{Lecture Notes in Computer Science}, vol. 1065, pp. 376--387. Springer,
  Berlin Heidelberg (1996)

\bibitem{Kubanek2012}
Kubanek, M., Bobulski, J., Adrjanowicz, L.: Characteristics of the use of
  coupled hidden markov models for audio-visual polish speech recognition.
\newblock Bull. of the Pol. Acad. of Sci.: Tech. Sci. \textbf{60}(2), 307--316
  (2012)

\bibitem{Lan2009}
Lan, Y., Harvey, R., Theobald, B., Ong, E.J., Bowden, R.: Comparing visual
  features for lipreading.
\newblock In: Proc. of the Int. Conf. on Auditory-Visual Speech Process., pp.
  102--106 (2009)

\bibitem{Lan2010}
Lan, Y., Theobald, B.J., Harvey, R., Ong, E.J., Bowden, R.: Improving visual
  features for lip-reading.
\newblock In: Proc. of the Int. Conf. on Auditory-Visual Speech Process., pp.
  142--147 (2010)

\bibitem{Li2009}
Li, M., Cheung, Y.M.: Automatic lip localization under face illumination with
  shadow consideration.
\newblock Signal Process. \textbf{89}(12), 2425--2434 (2009)

\bibitem{Liu2010}
Liu, X., Cheung, Y.M., Li, M., Liu, H.: A lip contour extraction method using
  localized active contour model with automatic parameter selection.
\newblock In: Proc. of the Int. Conf. on Pattern Recognition, pp. 4332--4335
  (2010)

\bibitem{Livescu2007}
Livescu, K., Cetin, O., Hasegawa-Johnson, M., King, S., Bartels, C., Borges,
  N., Kantor, A., Lal, P., Yung, L., Bezman, A., Dawson-Haggerty, S., Woods,
  B., Frankel, J., Magami-Doss, M., Saenko, K.: Articulatory feature-based
  methods for acoustic and audio-visual speech recognition: Summary from the
  2006 {JHU} summer workshop.
\newblock In: Proc. of the IEEE Int. Conf. on Acoust., Speech and Signal
  Process., vol.~4, pp. 621--624 (2007)

\bibitem{Mahdi2008}
Mahdi, W., Werda, S., Hamadou, A.B.: A hybrid approach for automatic lip
  localization and viseme classification to enhance visual speech recognition.
\newblock Integr. Computer-Aided Eng. \textbf{15}(3), 253--266 (2008)

\bibitem{Min2011}
Min, K.Y., Zuo, L.H.: A lip reading method based on {3-D DCT} and {3-D HMM}.
\newblock In: Proc. of the Int. Conf. on Electron. and Optoelectron., vol.~1,
  pp. 115--119 (2011)

\bibitem{Needleman1970}
Needleman, S.B., Wunsch, C.D.: A general method applicable to the search for
  similarities in the amino acid sequence of two proteins.
\newblock J. of Mol. Biol. \textbf{48}(3), 443 -- 453 (1970)

\bibitem{Neti2000}
Neti, C., Potamianos, G., Luettin, J., Matthews, I., Glotin, H., Vergyri, D.,
  Sison, J., Mashari, A., Zhou, J.: Audio-visual speech recognition.
\newblock In: Idiap Research Institute Final Workshop 2000 Report, vol. 764
  (2000)

\bibitem{Newman2010}
Newman, J.L., Theobald, B.J., Cox, S.J.: Limitations of visual speech
  recognition.
\newblock In: Proc. of Int. Conf. on Auditory-Visual Speech Process. (2010)

\bibitem{ONeill1951}
O'Neill, J.: An exploratory investigation of lipreading ability among normal
  hearing students.
\newblock Speech Monogr. \textbf{18}(4), 309--311 (1951)

\bibitem{Ong2008}
Ong, E.J., Bowden, R.: Robust lip-tracking using rigid flocks of selected
  linear predictors.
\newblock In: Proc. of the IEEE Int. Conf. on Autom. Face and Gesture
  Recognition (2008)

\bibitem{Ong2011}
Ong, E.J., Bowden, R.: Learning sequential patterns for lipreading.
\newblock In: Proc. of the 22nd Br. Mach. Vis. Conf., pp. 55--65 (2011)

\bibitem{Pandzic2003}
Pandzic, I.S., Forchheimer, R. (eds.): MPEG-4 Facial Animation: The Standard,
  Implementation and Applications.
\newblock John Wiley \& Sons, Inc., New York (2003)

\bibitem{Pass2010}
Pass, A., Zhang, J., Stewart, D.: An investigation into features for multi-view
  lipreading.
\newblock In: Proc. of the IEEE Int. Conf. on Image Process., pp. 2417--2420
  (2010)

\bibitem{Patterson2002}
Patterson, E.K., Gurbuz, S., Tufekci, Z., Gowdy, J.N.: {CUAVE}: A new
  audio-visual database for multimodal human-computer interface research.
\newblock In: Proc. of the IEEE Int. Conf. on Acoust., Speech, and Signal
  Process., vol.~2, pp. 2017--2020 (2002)

\bibitem{Petajan1984}
Petajan, E.D.: Automatic lipreading to enhance speech recognition (speech
  reading).
\newblock Ph.D. thesis, University of Illinois, Champaign, USA (1984)

\bibitem{Picard2010}
Picard, S., Ananthakrishnan, G., Wik, P., Engwall, O., Abdou, S.: Detection of
  specific mispronunciations using audiovisual features.
\newblock In: Proc. of the Int. Conf. on Auditory-Visual Speech Process. (2010)

\bibitem{Ramage2011}
Ramage, M.D.: Disproving visemes as the basic visual unit of speech.
\newblock Ph.D. thesis, Curtin University, Australia (2011)

\bibitem{Saenko2009}
Saenko, K., Livescu, K., Glass, J., Darrell, T.: Multistream articulatory
  feature-based models for visual speech recognition.
\newblock IEEE Trans. on Pattern Anal. and Mach. Intell. \textbf{31}(9),
  1700--1707 (2009)

\bibitem{Sanderson2009}
Sanderson, C., Lovell, B.C.: Multi-region probabilistic histograms for robust
  and scalable identity inference.
\newblock In: Proc. of the Int. Conf. on Adv. in Biom., ICB '09, pp. 199--208.
  Springer-Verlag, Berlin, Heidelberg (2009)

\bibitem{Sargin2007}
Sargın, M.E., Yemez, Y., Erzin, E., Tekalp, A.M.: Audiovisual synchronization
  and fusion using canonical correlation analysis.
\newblock IEEE Trans. on Multimed. \textbf{9}(7), 1396--1403 (2007)

\bibitem{Seyedarabi2006}
Seyedarabi, H., Lee, W., Aghagolzadeh, A.: Automatic lip tracking and action
  units classification using two-step active contours and probabilistic neural
  networks.
\newblock In: Proc. of the Can. Conf. on Electr. and Comput. Eng., pp.
  2021--2024 (2006)

\bibitem{Shoup1980}
Shoup, J.E.: Phonological aspects of speech recognition.
\newblock In: W.A. Lea (ed.) Trends in Speech Recognition, pp. 125--138.
  Prentice-Hall, Englewood Cliffs (1980)

\bibitem{Singh2012}
Singh, P., Laxmi, V., Gaur, M.S.: Relevant {mRMR} features for visual speech
  recognition.
\newblock In: Proc. of the Int. Conf. on Recent Adv. in Comput. and Softw.
  Syst., pp. 148--153 (2012)

\bibitem{Stillittano2013}
Stillittano, S., Girondel, V., Caplier, A.: Lip contour segmentation and
  tracking compliant with lip-reading application constraints.
\newblock Mach. Vision and Appl. \textbf{24}(1), 1--18 (2013)

\bibitem{Sujatha2010}
Sujatha, B., Santhanam, T.: A novel approach integrating geometric and gabor
  wavelet approaches to improvise visual lip-reading.
\newblock Int. J. of Soft Comput. \textbf{5}, 13--18 (2010)

\bibitem{Taylor2012}
Taylor, S.L., Mahler, M., Theobald, B.J., Matthews, I.: Dynamic units of visual
  speech.
\newblock In: Proc. of the ACM SIGGRAPH/Eurographics Symp. on Comput. Animat.,
  pp. 275--284. Aire-la-Ville (2012)

\bibitem{Tian2000}
Tian, Y.L., Kanade, T., Cohn, J.F.: Robust lip tracking by combining shape,
  color and motion.
\newblock In: Proc. of the 4th Asian Conf. on Comput. Vis., pp. 1040--1045
  (2000)

\bibitem{Viterbi1967}
Viterbi, A.: Error bounds for convolutional codes and an asymptotically optimum
  decoding algorithm.
\newblock IEEE Trans. on Inf. Theory \textbf{13}(2), 260--269 (1967)

\bibitem{Werda2007}
Werda, S., Mahdi, W., Hamadou, A.B.: Lip localization and viseme classification
  for visual speech recognition.
\newblock Int. J. of Comput. \& Inf. Sci. \textbf{5}(1), 62--75 (2007)

\bibitem{Yau2007}
Yau, W.C., Kumar, D.K., Weghorn, H.: Visual speech recognition using motion
  features and hidden markov models.
\newblock In: W.G. Kropatsch, M.~Kampel, A.~Hanbury (eds.) Computer Analysis of
  Images and Patterns, \emph{Lecture Notes in Computer Science}, vol. 4673, pp.
  832--839. Springer, Berlin Heidelberg (2007)

\bibitem{Yu2008}
Yu, D., Ghita, O., Sutherland, A., Whelan, P.F.: A novel visual speech
  representation and {HMM} classification for visual speech recognition.
\newblock In: Proc. of the 3rd Pac. Rim Symp. on Adv. in Image and Video
  Technol., pp. 398--409 (2008)

\bibitem{Zhao2009}
Zhao, G., Barnard, M., Pietik\"{a}inen, M.: Lipreading with local
  spatiotemporal descriptors.
\newblock IEEE Trans. on Multimed. \textbf{11}(7), 1254--1265 (2009)

\bibitem{Zue1990}
Zue, V., Seneff, S., Glass, J.: Speech database development at {MIT}: Timit and
  beyond.
\newblock Speech Commun. \textbf{9}(4), 351 -- 356 (1990)

\end{thebibliography}

\end{document}